\documentclass[letterpaper]{article} % DO NOT CHANGE THIS
\usepackage{acl}  % DO NOT CHANGE THIS

\usepackage{times} % DO NOT CHANGE THIS
\usepackage{latexsym}

\usepackage[T1]{fontenc}
\usepackage[utf8]{inputenc}

\usepackage{microtype}

\usepackage{inconsolata} % DO NOT CHANGE THIS

\usepackage{adjustbox}

\usepackage{graphicx} % DO NOT CHANGE THIS
\usepackage{xcolor}         
\usepackage{colortbl}         
\usepackage{amssymb} 

\usepackage{courier}  % DO NOT CHANGE THIS
\usepackage{natbib}  % DO NOT CHANGE THIS AND DO NOT ADD ANY OPTIONS TO IT
\usepackage{caption} % DO NOT CHANGE THIS AND DO NOT ADD ANY OPTIONS TO IT
\usepackage{algorithm}
\usepackage{algpseudocode}
\usepackage{bm}
\usepackage{subcaption}
\usepackage{booktabs}
\usepackage{multirow}
\usepackage{amsmath}
\usepackage{amsfonts}
\usepackage{cleveref}
\title{Normalized AOPC: Fixing Misleading Faithfulness Metrics for Feature Attribution Explainability}
\author{%
Joakim Edin$^{1,2*}$ \quad Andreas G. Motzfeldt $^{1,3*}$ \quad Casper L. Christensen$^{1*}$ \\
\textbf{Tuukka Ruotsalo}$^{2,4}$ \quad  \textbf{Lars Maaløe}$^1$ \quad \textbf{Maria Maistro}$^2$  \\
$^1$Corti \quad $^2$University of Copenhagen  \\$^3$IT University of Copenhagen \quad $^4$LUT University \\ $^*$Equal contributions  \\
\texttt{\{je,clu,amo\}@corti.ai}
}

\usepackage{bibentry}
\begin{document}

\maketitle

\begin{abstract}
Deep neural network predictions are notoriously difficult to interpret. Feature attribution methods aim to explain these predictions by identifying the contribution of each input feature. Faithfulness, often evaluated using the area over the perturbation curve (AOPC), reflects feature attributions' accuracy in describing the internal mechanisms of deep neural networks. However, many studies rely on AOPC to compare faithfulness across different models, which we show can lead to false conclusions about models' faithfulness. Specifically, we find that AOPC is sensitive to variations in the model, resulting in unreliable cross-model comparisons. Moreover, AOPC scores are difficult to interpret in isolation without knowing the model-specific lower and upper limits. To address these issues, we propose a normalization approach, Normalized AOPC (NAOPC), enabling consistent cross-model evaluations and more meaningful interpretation of individual scores. Our experiments demonstrate that this normalization can radically change AOPC results, questioning the conclusions of earlier studies and offering a more robust framework for assessing feature attribution faithfulness. Our code is available at \url{https://github.com/JoakimEdin/naopc}.
\end{abstract}

\section{Introduction}

Deep neural networks are often described as black boxes due to the difficulty in understanding their inner mechanisms~\cite{wei2022emergent}. This lack of interpretability can hinder their adoption in critical applications where trust is paramount~\cite{lipton2018mythos}. For instance, if a diagnostic model predicts meningitis without an explanation, a physician confident in an influenza diagnosis might incorrectly dismiss the model's prediction as an error.

Feature attribution methods attempt to address this issue by quantifying each input feature's contribution to a model's output~\cite{danilevskySurveyStateExplainable2020}. In the meningitis example, such a method might identify ``fever'' and ``stiff neck'' as important features, potentially convincing the physician to reconsider the diagnosis. For these methods to be reliable, they must faithfully represent the model's underlying reasoning process, avoiding misleading interpretations~\cite{jacoviFaithfullyInterpretableNLP2020}.

The Area Over the Perturbation Curve (AOPC) has become a standard metric for approximating faithfulness, with two main variants: sufficiency and comprehensiveness~\cite{deyoungERASERBenchmarkEvaluate2020, lyu2024faithfulmodelexplanationnlp}. However, we uncover a critical weakness: the minimum and maximum possible AOPC scores vary significantly across different models and inputs. For the same dataset, we found one model's upper limit averaged 0.3 across examples while another averaged 0.8. These varying lower and upper limits of AOPC stem from how models transform inputs into outputs, specifically, how many input features each model uses and how it combines these features through interactions to produce predictions. 

This weakness invalidates AOPC comparisons across models, affecting several studies in explainable AI. We found eleven studies in top machine learning venues that compare AOPC across models. These studies are spread among the following research directions: learning to explain~\cite{resckExploringTradeModelPerformance2024, liuImproveInterpretabilityNeural2022}, developing self-explanatory model architectures~\cite{sekhonImprovingInterpretabilityExplicit2023}, making models more interpretable through training strategies (e.g., adversarial robustness)~\cite{bhallaDiscriminativeFeatureAttributions2023,liFaithfulExplanationsText2023, chenLearningVariationalWord2020, chrysostomouEnjoySalienceBetter2021, xieIvRAFrameworkEnhance2024}, and analyzing explanation methods in different settings (e.g., out-of-distribution data)~\cite{chrysostomouEmpiricalStudyExplanations2022, haseOutofDistributionProblemExplainability2021, nielsenEvalAttAIHolisticApproach2023}. Our findings suggest these studies' findings may be misguiding.

The varying limits also make AOPC scores difficult to interpret. For instance, is an AOPC score of 0.25 high? It is high if the upper limit is 0.3 but not if it is 0.8. Interpretable scores would help researchers identify models and inputs where explanation methods produce unfaithful explanations, enabling them to analyze these cases and debug their methods. Without knowing the model and input-specific limits, this systematic improvement of explanation methods remains difficult.

To address these issues, we propose Normalized AOPC (NAOPC), an approach to normalize AOPC scores to ensure comparable lower and upper limits across all models and data examples. Our empirical results show that this normalization can significantly alter the faithfulness ranking of models, questioning previous conclusions about improved model faithfulness.
Our key contributions are:

\begin{enumerate}
    \item We demonstrate that the minimum and maximum possible AOPC scores vary significantly across different models and inputs, which makes cross-model comparisons and isolated score interpretations problematic. 
    \item We propose NAOPC, including an exact version (NAOPC$_{\text{exact}}$) and a faster approximation (NAOPC$_{\text{beam}}$), to normalize AOPC scores for improved comparability. 
    \item We show empirically, with five datasets, four model architectures, and three NLP tasks, how NAOPC alters faithfulness rankings, highlighting the need to re-evaluate conclusions in previous studies about model faithfulness.
\end{enumerate}

To facilitate adoption of these methods, we release AOPC, NAOPC$_{\text{exact}}$, and NAOPC$_{\text{beam}}$ as a PyPI package\footnote{Our names are mentioned in the PyPI package. We will include a link in the camera-ready version.}.

% \footnote{Sufficiency is often referred to as \textit{Least Relevant First},  comprehensiveness as \textit{Most Relevant first}, and collectively they are referred to as \textit{fidelity}~\cite{wangInterpretingInterpretationsOrganizing2020,carton2020evaluating}.}

\section{Problem formulation}
\noindent\textbf{Area Over the Perturbation Curve (AOPC)} measures the change in model output as input features are sequentially perturbed~\cite{samek2016evaluating}. The perturbation can either remove, insert, or replace a feature with some pre-defined value. The final score is the average output change across all perturbation steps. Formally, the AOPC is calculated as follows:
\begin{equation}
    \text{AOPC}(f, \bm{x}, \bm{r}) = \frac{1}{N} \sum_{i=1}^{N} f(\bm{x})-f(p(\bm{x}, \bm{r}_{1:i}))
\end{equation}
where $f$ is a model, $\bm{x}$ is an input vector with $N$ number of features, $\bm{r} \in \text{Permutations}(\{1, \ldots, N\})$ is the order to perturb the input features, and $p$, is the perturbation function that removes, inserts or replaces the features in $\bm{x}$ that are in $\bm{r}_{1:i}$. AOPC is used to calculate sufficiency and comprehensiveness as follows. \\

\noindent\textbf{Comprehensiveness} estimates the faithfulness of feature attribution scores by perturbing the input features in decreasing order, starting from the feature with the highest score~\cite{deyoungERASERBenchmarkEvaluate2020}. A high comprehensiveness indicates that the features that are the highest ranked, according to the feature attributions, are important for the model's output (i.e., higher is better). Comprehensiveness is calculated as follows:
\begin{equation}
    \text{Comp}(f, \bm{x}, \bm{e}) = \text{AOPC}(f, \bm{x}, \text{rank}(\bm{e}))
\end{equation}
\noindent where $\text{rank}(\cdot)$ returns the ordering of the feature attribution scores $\bm{e}$ in decreasing order. \\

\noindent\textbf{Sufficiency} perturbs the input features in increasing order, starting from the feature with the lowest score~\cite{deyoungERASERBenchmarkEvaluate2020}. In other words, sufficiency is comprehensiveness when flipping the feature ordering $r$. A low sufficiency indicates that the lowest ranked features, according to the feature attributions, are irrelevant to the model output (i.e., lower is better). Sufficiency is calculated as follows:
\begin{equation}
    \text{Suff}(f, \bm{x}, \bm{e}) = \text{AOPC}(f, \bm{x}, \text{rank}(-\bm{e}))
\end{equation}
Notably, the best possible sufficiency and comprehensiveness scores correspond to the empirical lower and upper limits of AOPC scores, respectively. Sufficiency seeks the feature ordering that produces the lowest possible AOPC score, whereas comprehensiveness aims for the ordering that yields the highest score. The best (lowest) possible sufficiency score is the worst (lowest) possible comprehensiveness score, and vice versa.

\subsection{Models influence AOPC scores}

Ideally, sufficiency and comprehensiveness should only measure the feature attribution's faithfulness. However, in this section, we demonstrate that a model's reasoning process heavily influences these AOPC metrics. Specifically, we show with two toy examples that 1) the more features a model relies on, the worse the sufficiency and comprehensiveness scores, and 2) a model's features interactions impact the best possible comprehensiveness and sufficiency scores. \\

\noindent\textbf{The number of features models rely on impact AOPC scores}. We demonstrate this with two linear models $f_1$ and $f_2$ that take four binary features as input $\bm{x}=(x_1, x_2, x_3, x_4)$ and output a real number:
\begin{equation}
    f_1(\bm{x}) = 0.2x_1 + 0.3x_2 + 0.1x_3 + 0.4x_4
\end{equation}
\begin{equation}
    f_2(\bm{x}) = 0.0x_1 + 0.1x_2 + 0.7x_3 + 0.2x_4
\end{equation}
The two models use the same architecture, but their parameter values differ. $f_2$ relies heavily on feature $x_3$, while $f_1$ relies on more features. Different training strategies, data, or randomness can cause such model differences~\cite{haseOutofDistributionProblemExplainability2021}. 

Given an input vector $\bm{x^{(0)}}=(1,1,1,1)$, both models output $1.0$. Since the models are linear, we can calculate the ground truth feature attributions by multiplying each input feature by its parameter. For $\bm{x^{(0)}}$, this results in $(0.2, 0.3, 0.1, 0.4)$ for $f_1$, and $(0.0, 0.1, 0.7, 0.2)$ for $f_2$. As these attributions represent the ground truths, one might expect the models to yield equal sufficiency and comprehensiveness scores. However, this is not the case. As shown in \Cref{tab:f1f2}, $f_1$ achieves drastically worse sufficiency and comprehensiveness scores than $f_2$ simply because it relies on more features. While relying on fewer features could be a desirable model property, it should be measured using entropy instead of influencing the faithfulness evaluation~\cite{bhattEvaluatingAggregatingFeaturebased2020}.
\begin{table}[ht]
    \caption{The comprehensiveness and sufficiency scores calculated given input $\bm{x}^{(0)}$ and the ground truth feature attribution scores for $f_1$ and $f_2$. Despite the feature attribution method being perfectly faithful, the comprehensiveness and sufficiency scores are better for model $f_2$ because it relies on fewer input features than $f_1$.}
    \label{tab:f1f2}
    \adjustbox{max width=\textwidth}{%
    \centering
    \begin{tabular}{ccc}
    \toprule
         Model & Comprehensiveness $\uparrow$& Sufficiency $\downarrow$ \\
         \midrule
         $f_1$ & 0.75 & 0.50 \\
         $f_2$ & \textbf{0.90} & \textbf{0.35} \\
         \bottomrule
    \end{tabular}
    }
\end{table}
\\
\noindent\textbf{Feature interactions in the models' reasoning process impact the AOPC scores}. We demonstrate this with two nonlinear models, $f_3$ and $f_4$, which use logical operations between the input features to generate the output score. $f_3$ uses OR-gates, while $f_4$ uses AND-gates
\begin{equation}
    f_3(\bm{x}) = 0.7 (x_1 \lor x_2) + 0.3(x_3 \lor x_4)
\end{equation}
\begin{equation}
    f_4(\bm{x}) = 0.7 (x_1 \land x_2) + 0.3(x_3 \land x_4)
\end{equation}
We cannot calculate the feature attribution scores for these models by multiplying the input features with their parameters as they are nonlinear. Instead, we use an exhaustive search algorithm to find the best comprehensiveness and sufficiency scores. This algorithm evaluates all possible feature orderings $r$ and identifies the highest and lowest scores. 

In \Cref{tab:f3f4}, we show the best sufficiency and comprehensiveness scores for $f_3$ and $f_4$ when given the input $\bm{x}^{(0)}$. $f_3$ achieves the best sufficiency score, while $f_4$ achieves the best comprehensiveness score. This indicates that the type of feature interactions a model uses impacts the best possible comprehensiveness and sufficiency scores. 

We expect these findings to extend to deep neural networks as well. In these more complex models, various components, such as activation functions and attention mechanisms, induce feature interactions~\cite{tsangHowDoesThis2020}. 

\begin{table}[ht]
    \caption{The best possible comprehensiveness and sufficiency scores for the two models $f_3$ and $f_4$ when given input $\bm{x}^{(0)}$. The models' feature interaction differences cause different scores.}
    \label{tab:f3f4}
    \adjustbox{max width=\textwidth}{%
    \centering
    \begin{tabular}{ccc}
    \toprule
         Model & Comprehensiveness $\uparrow$& Sufficiency $\downarrow$ \\
         \midrule
         $f_3$ & 0.6 & \textbf{0.325} \\
         $f_4$ & \textbf{0.925} & 0.65 \\
         \bottomrule
    \end{tabular}
    }
\end{table}

Recall that the best possible sufficiency and comprehensiveness scores correspond to the lower and upper limits of AOPC scores. Because we have shown that the four models' best possible sufficiency and comprehensiveness scores vary for the same input, we have also shown that they have different lower and upper limits of AOPC scores. Consequently, we have demonstrated that given input $\bm{x^{(0)}}$, feature attribution methods can only achieve AOPC scores between 0.5--0.75 for $f_1$, 0.35--0.9 for $f_2$, 0.325--0.6 for $f_3$, and 0.65--0.925 for $f_4$, which makes the models' scores uncomparable. In the next section, we will propose methods for normalizing AOPC so that all models have the same lower and upper limits, making them comparable.

\section{Normalized AOPC}
Our previous analysis revealed that AOPC limits can vary between models for the same input, even for linear models. To address this issue, we propose Normalized AOPC (NAOPC), which ensures comparable AOPC scores across different models and inputs. NAOPC applies min-max normalization to the AOPC scores using their lower and upper limits:
\begin{equation}
\text{NAOPC}(f, \bm{x}, \bm{r}) = \frac{\text{AOPC}(f, \bm{x}, \bm{r})-\text{AOPC}_{\uparrow}(f, \bm{x})}{\text{AOPC}_{\uparrow}(f, \bm{x})-\text{AOPC}_{\downarrow}(f, \bm{x})}
\end{equation}
\noindent where $\text{AOPC}_{\downarrow}(f, \bm{x})$ and $\text{AOPC}_{\uparrow}(f, \bm{x})$ represent the lower and upper AOPC limits for a specific model $f$ and input $\bm{x}$. We propose two variants of NAOPC, differing in how they identify these limits:

\paragraph{NAOPC$_{\text{exact}}$} uses an exhaustive search to find the exact lower and upper AOPC limits. It calculates the AOPC score for all $N!$ possible feature orderings $\bm{r}$, where $N$ is the number of features. While precise, its $O(N!)$ time complexity makes it prohibitively slow for high-dimensional inputs.

\paragraph{NAOPC$_{\text{beam}}$} efficiently approximates NAOPC$_{\text{exact}}$ using beam search to find the lower and upper AOPC limits (See \Cref{alg:naopc_beam}). Inspired by \citet{zhou2022solvability}, it runs twice: once for each limit. In each run, it maintains a beam of the top $B$ feature orderings, expanding them incrementally until all features are ordered. This approach limits the search space, achieving a time complexity of $O(B \cdot N^2)$, where $B$ is the beam size and $N$ is the number of features. Consequently, NAOPC$_{\text{beam}}$ is significantly faster than NAOPC$_{\text{exact}}$ for high-dimensional inputs while still providing a good approximation of the AOPC limits.

To select an appropriate beam size for NAOPC, we check if the upper and lower AOPC limits remain stable as we increase the beam size~\cite{freitagBeamSearchStrategies2017}. We start with a beam size of 1 and double the beam size until the limits do not change more than a pre-selected threshold two iterations in a row.

\begin{algorithm}[h]
\caption{NAOPC$_{\text{beam}}$}
\label{alg:naopc_beam}
\begin{algorithmic}[1]
\Require Model $f$, input $\bm{x}$, beam size $B$, ordering $\bm{r}$
\Ensure Normalized AOPC score
\Function{FindLimit}{$f$, $\bm{x}$, $B$, $\text{mode}$}
    \State $\text{fullOutput} \gets f(\bm{x})$
    \State $\text{beam} \gets \{[]\}$
    \State $\text{scores} \gets \{(): 0\}$
    \For{$i = 1$ to $N$}
        \State $\text{cand} \gets \{\}$
        \For{$\text{ord} \in \text{beam}$}
            \For{$j \in \{1,\ldots,N\} \setminus \text{ord}$}
                \State $\text{new\_ord} \gets \text{ord} + [j]$
                \State $\bm{\hat{x}}\gets \text{MaskTokens}(\bm{x}, \text{new\_ord})$
                \State $\text{score} \gets \text{fullOutput} - f(\bm{\hat{x}})$
                \State $\text{score} \gets \text{score} + \text{scores}[\text{ord}]$
                \State $\text{scores}[\text{new\_ord}] \gets \text{score}$
                \State $\text{cand} \gets \text{cand} \cup \{(\text{new\_ord}, \text{score})\}$
            \EndFor
        \EndFor
        \If{$\text{mode} = \text{``upper''}$}
            \State $\text{beam} \gets \text{TopB}(\text{cand}, B, \text{max})$
        \Else
            \State $\text{beam} \gets \text{TopB}(\text{cand}, B, \text{min})$
        \EndIf
    \EndFor
    \State $\text{AOPC} \gets \frac{\text{beam}[0]}{N}$
    \State \Return $\text{AOPC}$
\EndFunction
\State $\text{upper\_limit} \gets \text{FindLimit}(f, \bm{x}, B, \text{``upper''})$
\State $\text{lower\_limit} \gets \text{FindLimit}(f, \bm{x}, B, \text{``lower''})$
\State $\text{aopc\_score} \gets \text{AOPC}(f, \bm{x}, \bm{r})$ \Comment{$\bm{r}$ is the original feature attribution ordering}
\State \Return $\frac{\text{aopc\_score} - \text{lower\_limit}}{\text{upper\_limit} - \text{lower\_limit}}$
\end{algorithmic}
\end{algorithm}
\section{Experimental Setup}
Our experiments address three key questions: 1) Do AOPC lower and upper limits vary across deep neural network models? 2) How does NAOPC affect model faithfulness rankings? and 3) How accurately does NAOPC$_{\text{beam}}$ approximate NAOPC$_{\text{exact}}$? This section outlines the experimental designs, including the datasets, models, and feature attribution methods employed to investigate these questions. 

\paragraph{Data}
Our experiments use five datasets: Yelp, IMDB, SST2, AG-News, and SNLI. Yelp, IMDB, and SST2 are sentiment classification datasets, which we chose for their prevalence in cross-model AOPC score comparison studies~\cite{haseOutofDistributionProblemExplainability2021, bhallaDiscriminativeFeatureAttributions2023, liFaithfulExplanationsText2023}. AG-News is a text classification dataset, and SNLI is a natural language inference dataset~\cite{zhangCharacterlevelConvolutionalNetworks2015, maccartneyModelingSemanticContainment2008}. We include these two datasets to evaluate whether our findings generalize to other tasks than sentiment classification. 

To address varying computational requirements and ensure comprehensive analysis, we create short-sequence and long-sequence subsets from each dataset's test set\footnote{For SST2, we use the validation set as our test set, as its test set is unlabeled.}. \Cref{tab:data_stats} summarizes the key statistics of our subsets. The short subsets, Yelp$_{\text{short}}$ and SST2$_{\text{short}}$, contain examples with up to 12 features, enabling computationally intensive evaluations such as NAOPC$_{\text{exact}}$. 

We create five long-sequence subsets: SST2$_{\text{long}}$, Yelp$_{\text{long}}$, IMDB$_{\text{long}}$, AG-News$_{\text{long}}$, and SNLI$_{\text{long}}$. We randomly sample 1000 examples from each dataset except SST2 (SST2 only comprises 400 examples). We choose this sample size to balance computational feasibility with the need for a statistically significant sample size. We exclude examples exceeding 512 tokens due to model constraints. 
 \begin{table}[ht]
    \centering
    \caption{Summary statistics of the dataset subsets used in this study. The number of words per example is presented as the median and IQR.}
    \label{tab:data_stats}
    \adjustbox{max width=\textwidth}{%
    \begin{tabular}{ccc}
    \toprule
         & \# examples & \# words per example\\
         \midrule
       Yelp$_{\text{short}}$  & 339 & 5 (4--7)\\
       SST2$_{\text{short}}$  & 66 & 8 (7--9)\\
       \midrule
       Yelp$_{\text{long}}$  & 1000 & 52 (30--92)\\
       SST2$_{\text{long}}$  & 400 & 19 (13--26)\\
       IMDB$_{\text{long}}$  & 1000 & 132 (98--173)\\
       AG-News$_{\text{long}}$  & 1000 & 37 (31--42)\\
       SNLI$_{\text{long}}$  & 1000 & 20 (16--26)\\
       \bottomrule
    \end{tabular}
    }
\end{table}

\begin{figure*}[h]
    \centering
    \begin{subfigure}{0.37\textwidth}
        \includegraphics[width=\linewidth]
        {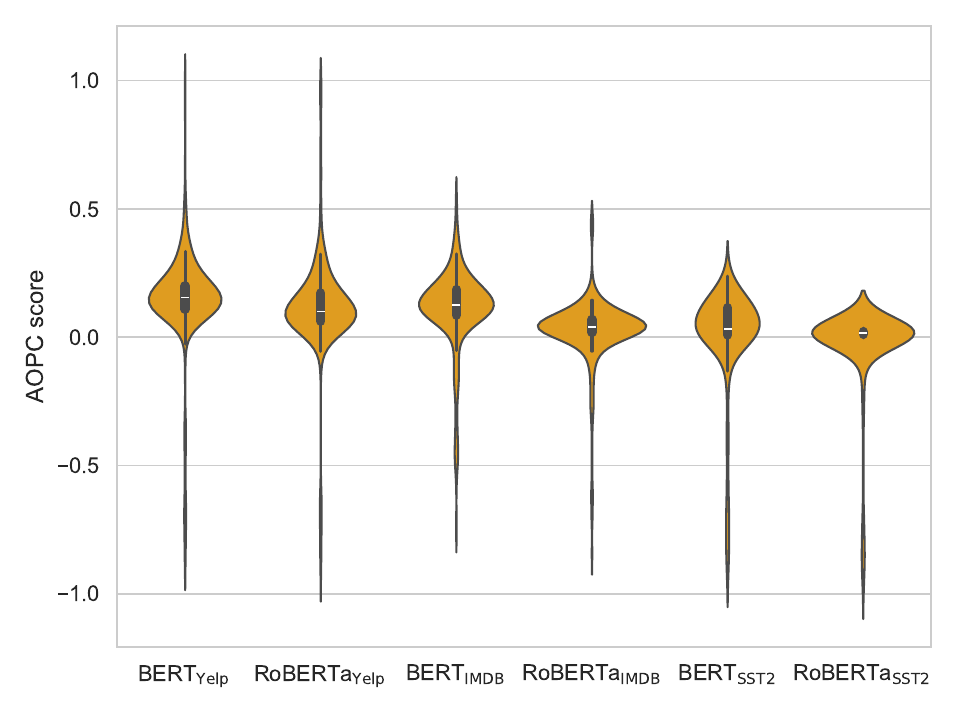}
        \caption{Lower limit}
    \end{subfigure}
    \begin{subfigure}{0.37\textwidth}
        \includegraphics[width=\linewidth]{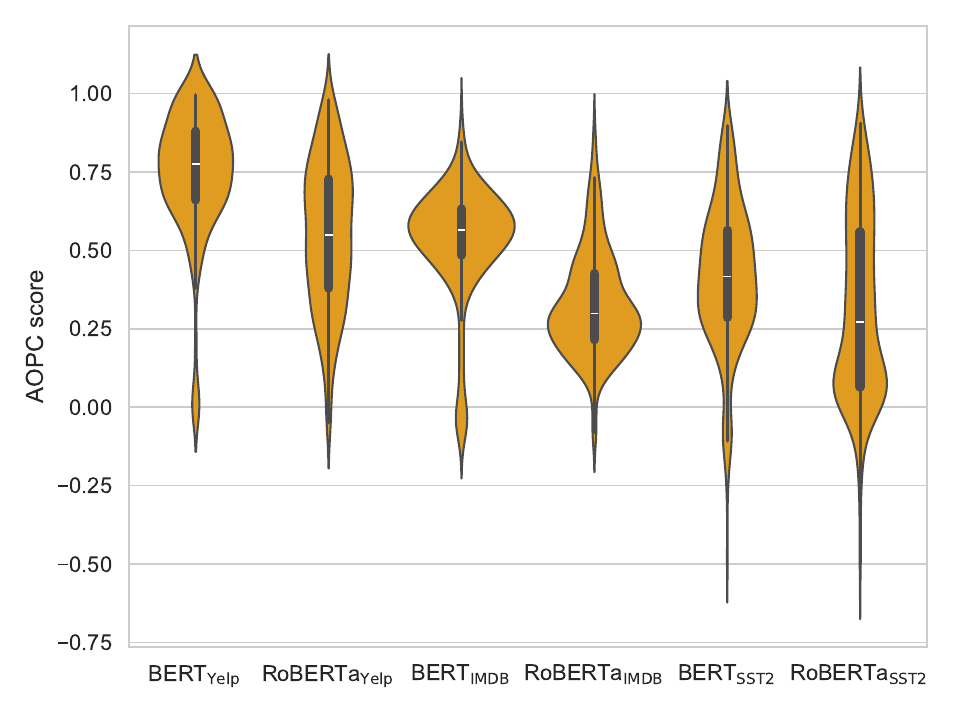}
        \caption{Upper limit}
    \end{subfigure}

    \caption{Distributions of lower and upper AOPC limits across models on the Yelp$_{\text{short}}$ test set computed with exhaustive search. The substantially different distributions demonstrate that AOPC bounds are model-specific, making both cross-model comparisons and interpretation of individual scores unreliable without normalization.}
    \label{fig:upper-lower}
\end{figure*}

\begin{figure*}[ht]
    \centering
    \includegraphics[width=0.75\textwidth]{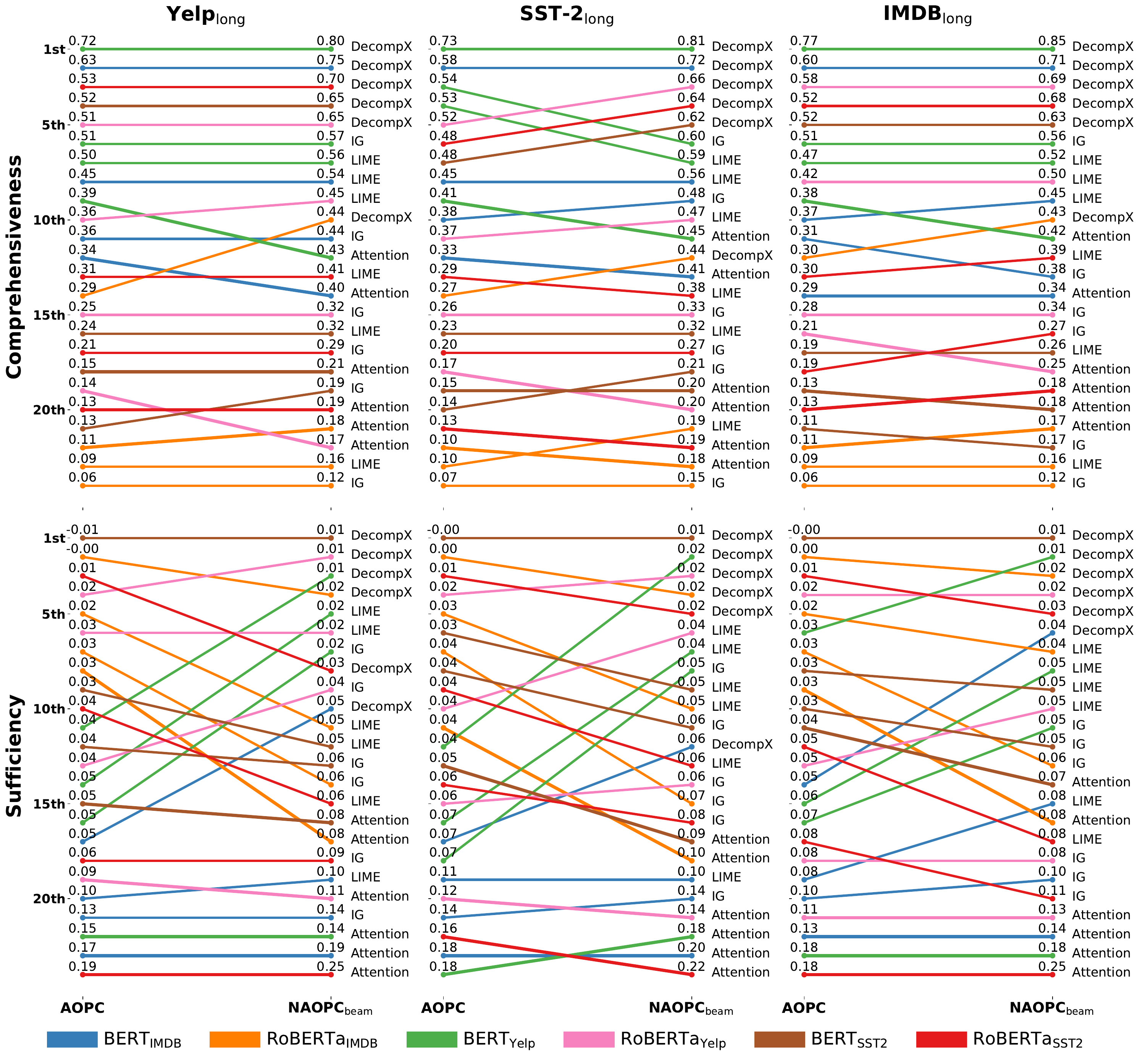}
    \caption{Effect of normalization on faithfulness rankings across models and attribution methods. For both comprehensiveness (higher is better) and sufficiency (lower is better), NAOPC$_{\text{beam}}$ changes cross-model rankings but preserves within-model rankings.}
    \label{fig:normalization_results_long}
\end{figure*}

\paragraph{Models}
Our experiments use twelve language models, all publicly available on Huggingface ~\cite{morrisTextAttackFrameworkAdversarial2020}. These models differ in two key aspects:

\begin{enumerate}
    \item \textbf{Architecture:} Each model is based on either BERT~\cite{devlinBERTPretrainingDeep2019}, DistilBERT~\cite{sanhDistilBERTDistilledVersion2020}, RoBERTa~\cite{liuRoBERTaRobustlyOptimized2019}, or GPT-2~\cite{radfordLanguageModelsAre} allowing us to examine how architectural differences influence AOPC scores and their interpretation.
    \item \textbf{Training Dataset:} Each model was trained on either Yelp, IMDB, SST2, AG-News, or SNLI~\cite{zhangCharacterlevelConvolutionalNetworks2015, maasLearningWordVectors2011, socherRecursiveDeepModels2013}, enabling investigation of how dataset-specific characteristics affect AOPC score limits. The lowercase suffix in each model's name (e.g., BERT$_{\text{Yelp}}$) indicates the dataset on which it was trained.
\end{enumerate}

We provide an overview of the models in \Cref{tab:apdx_models}.

\paragraph{Feature Attribution Methods} 
We implement eight feature attribution methods: two transformer-specific, three gradient-based, and three perturbation-based (see \citet{lyu2024faithfulmodelexplanationnlp} for an extensive overview of feature attribution methods). The \textit{transformer-specific} methods, Attention~\cite{jainAttentionNotExplanation2019} and DecompX~\cite{modarressiDecompXExplainingTransformers2023a} are specifically designed for transformer architectures. Attention calculates the feature attribution scores only using the attention weights in the final layer, while DecompX uses all the components and layers in the transformer architecture. The \textit{gradient-based} methods, InputXGrad~\cite{sundararajanAxiomaticAttributionDeep2017a}, Integrated Gradients~\cite{sundararajanAxiomaticAttributionDeep2017a}, and Deeplift~\cite{shrikumarLearningImportantFeatures2017}, use backpropagation to quantify the influence of input features on output. The \textit{perturbation-based} methods, LIME~\cite{ribeiroWhyShouldTrust2016}, KernelSHAP~\cite{lundbergUnifiedApproachInterpreting2017}, and Occlusion@1~\cite{ribeiroWhyShouldTrust2016}, assess the impact on output confidence by occluding input features.

In this paper, we use a perturbation function that replaces tokens with the mask token to calculate IG, Deeplift, LIME, KernelSHAP, and Occlusion@1. We also use this perturbation function to calculate the AOPC scores as recommended by \citet{haseOutofDistributionProblemExplainability2021}. We used the end-of-sequence token for GPT-2 because it does not support mask tokens nor pad tokens.

\paragraph{Experiment 1: Do the upper and lower limits vary between models?}
We aim to show that the lower and upper limits of the AOPC scores vary between the models and inputs. We do so by calculating each model's upper and lower AOPC limits using an exhaustive search for each example in Yelp$_{\text{short}}$ (same search strategy used by NAOPC$_{\text{exact}}$). We then compare the models' lower and upper limit distributions to demonstrate their differences. 

\paragraph{Experiment 2: How does normalization impact AOPC scores?}
In this experiment, we aim to answer the following two questions:
\begin{enumerate}
    \item Can NAOPC alter the faithfulness ranking of models for a given feature attribution method?
    \item Can NAOPC alter the faithfulness ranking of feature attribution methods for a given model?
\end{enumerate}
To answer these questions, we compare the sufficiency and comprehensiveness scores using AOPC and NAOPC. Specifically, we compare AOPC with NAOPC$_{\text{beam}}$ for all possible pairs of models and feature attribution methods on the long-sequence datasets. For the short-sequence datasets, we compare AOPC with both NAOPC$_{\text{exact}}$ and NAOPC$_{\text{beam}}$. We analyze the results to find whether NAOPC changes the ranking of which models and feature attribution methods are the most faithful. We use a beam size of 5 when calculating NAOPC$_{\text{beam}}$ on all datasets except for AG-News, which required a beam size of 1000.

\paragraph{Experiment 3: Can we approximate NAOPC$_{\text{exact}}$ reliably and efficiently?} We aim to demonstrate that NAOPC$_{\text{beam}}$ is a fast and reliable approximation of NAOPC$_{\text{exact}}$. With a sufficiently large beam size, NAOPC$_{\text{beam}}$ is equivalent to NAOPC$_{\text{exact}}$. The question is if NAOPC$_{\text{beam}}$ can accurately approximate NAOPC$_{\text{exact}}$ with small beam sizes.

First, we demonstrate that the faithfulness rankings produced with NAOPC$_{\text{beam}}$ and NAOPC$_{\text{exact}}$ are similar on Yelp$_{\text{short}}$ and SST2$_{\text{short}}$ with a beam size of 5. We cannot make this comparison on the five long-sequence dataset subsets because calculating NAOPC$_{\text{exact}}$ on high-dimensional inputs is prohibitively slow. Instead, we calculate NAOPC$_{\text{beam}}$ with increasing beam sizes and analyze the change of the lower and upper AOPC limits. If the AOPC limits stabilize at small beam sizes, it indicates that NAOPC$_{\text{beam}}$ can efficiently and reliably approximate NAOPC$_{\text{exact}}$.

\section{Results}

\subsection{The lower and upper limits vary between models}

\Cref{fig:upper-lower} shows significant variations in the distributions of lower and upper AOPC score limits across different models on the Yelp$_{\text{short}}$ test set. Each model has a distribution rather than a single value because individual inputs also influence the AOPC limits for each model. The clear differences in these distributions across models highlight that direct comparisons of AOPC scores between models can be misleading without proper normalization. Moreover, these variations make interpreting AOPC scores in absolute terms challenging. Figure \ref{fig:upper-lower} depicts an upper limit of around 0.3 for RoBERTa$_{\text{IMDB}}$ and 0.8 for BERT$_{\text{Yelp}}$, therefore an AOPC score of 0.25 might be considered high for RoBERTa$_{\text{IMDB}}$ but low for BERT$_{\text{Yelp}}$. These distribution shifts across models emphasize the need for NAOPC. \Cref{fig:sst2-upper-lower} depicts a similar pattern on SST2$_{\text{short}}$.

\subsection{NAOPC alters faithfulness rankings}
\begin{figure*}[ht]
    \centering
    \begin{subfigure}{0.37\textwidth}
        \includegraphics[width=\textwidth]{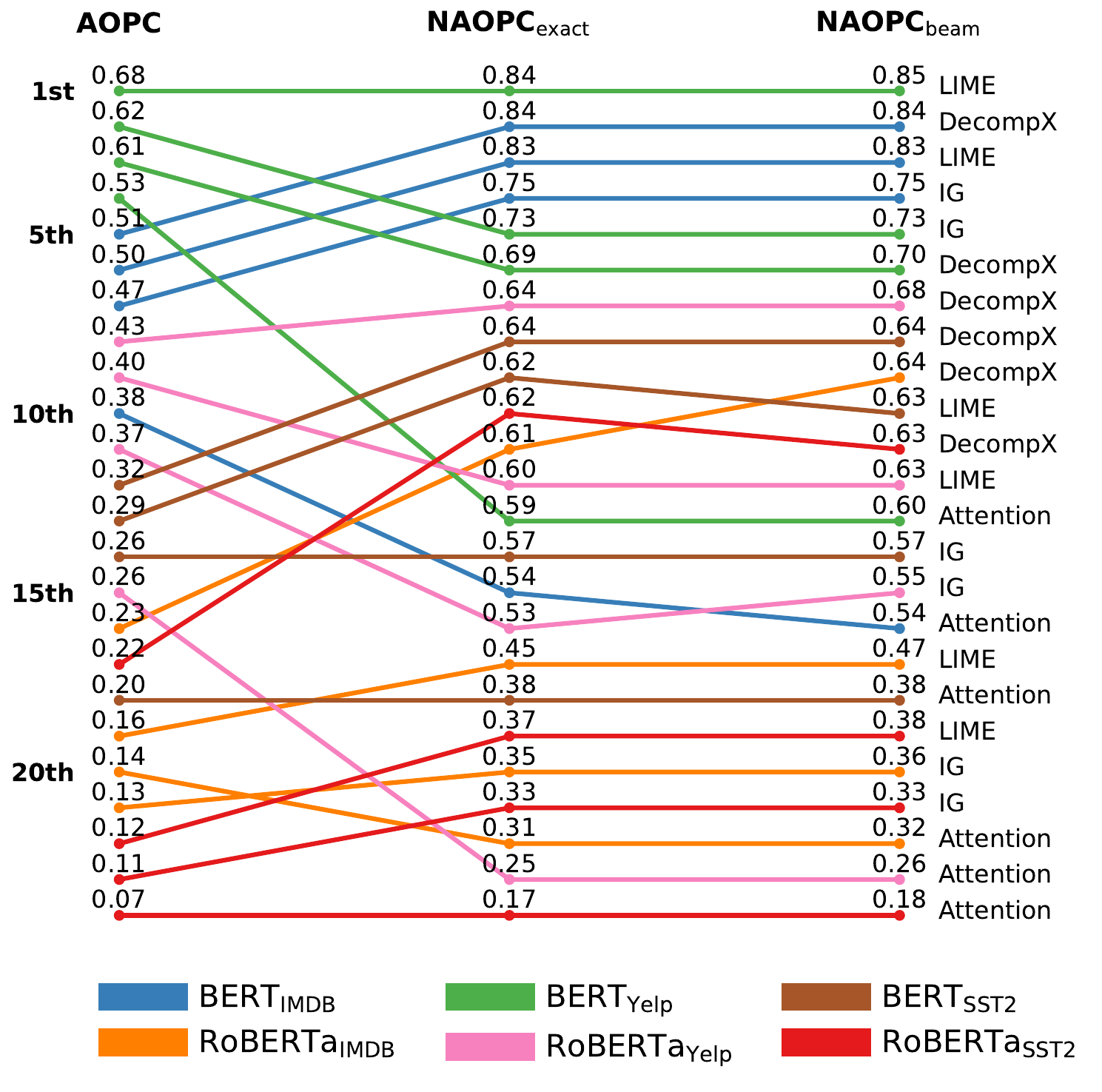}
        \caption{Comprehensiveness ranking}
    \end{subfigure}
    \begin{subfigure}{0.37\textwidth}
        \includegraphics[width=\textwidth]{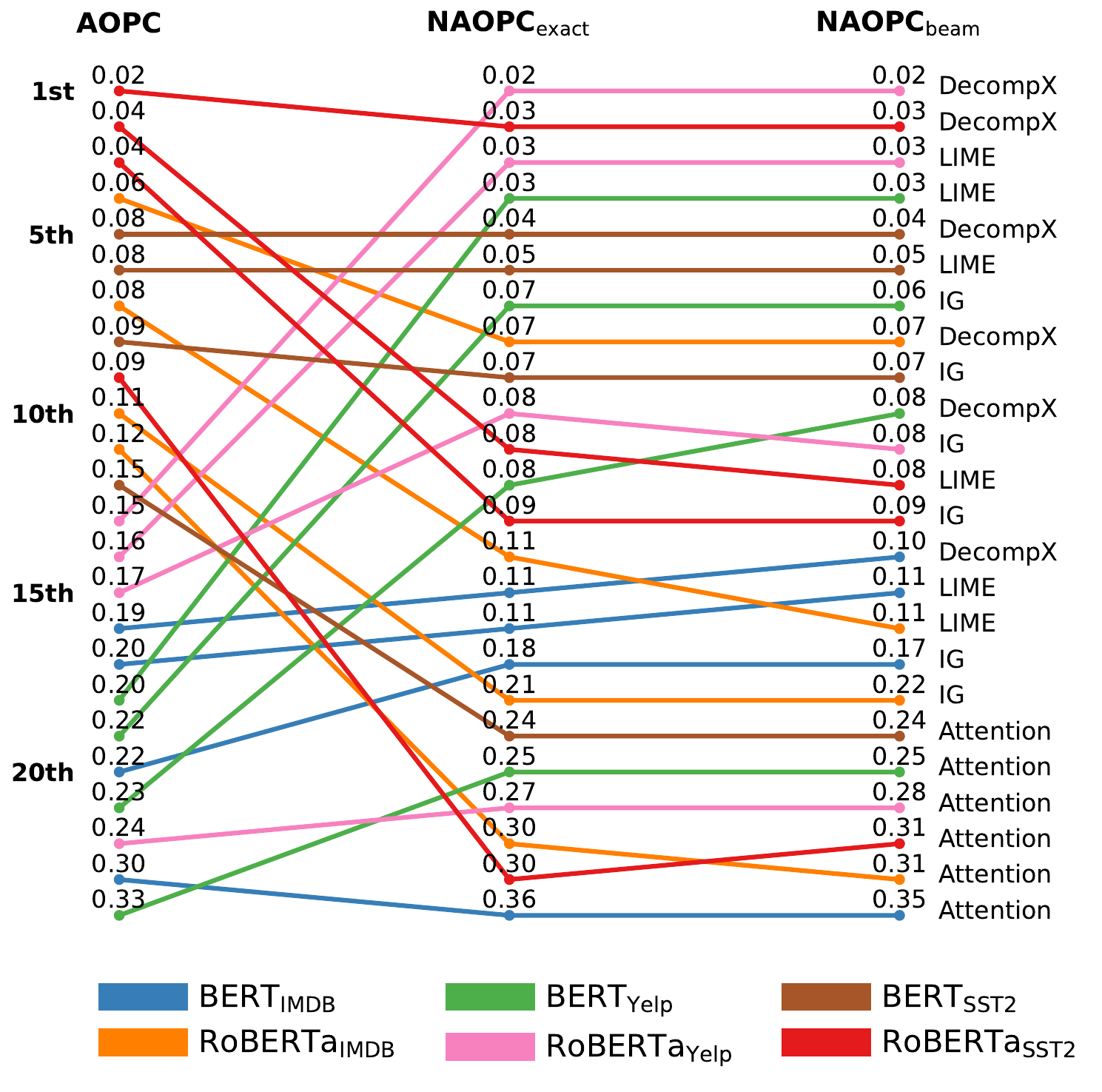}
        \caption{Sufficiency ranking}
    \end{subfigure}

    \caption{Faithfulness ranking of model and feature attribution method pairs when evaluated on Yelp$_{\text{short}}$ using AOPC, NAOPC$_{\text{exact}}$, and NAOPC$_{\text{beam}}$. The figure shows that normalization changes the cross-model comparisons and that NAOPC$_{\text{beam}}$ accurately approximates NAOPC$_{\text{exact}}$}
    \label{fig:normalization_results_short}
\end{figure*}

Normalization through NAOPC substantially altered models' faithfulness rankings while preserving the relative performance of feature attribution methods. This effect is clearly visible in \Cref{fig:normalization_results_long}, where lines of different colors (representing different models) frequently intersect between AOPC and NAOPC$_{\text{beam}}$ rankings across Yelp$_{\text{long}}$, IMDB$_{\text{long}}$, and SST2$_{\text{long}}$ datasets. In contrast, lines of the same color (representing feature attribution methods within a model) rarely cross, indicating stability in their relative rankings. We see a similar trend on AG-News$_{\text{long}}$ and SNLI$_{\text{long}}$ in \Cref{fig:normalization_results_agnews_snli}. The impact of normalization appears even more pronounced in shorter text datasets, as shown in \Cref{fig:normalization_results_short} for Yelp$_{\text{short}}$. 

These visual observations are quantitatively supported by the Kendall rank correlation coefficients presented in \Cref{tab:correlations_all}~\cite{kendall1948rank}. Correlations between AOPC and NAOPC$_{\text{beam}}$ scores are notably lower for the model comparisons than for the feature attribution comparisons. This pattern is consistent across all datasets.
\begin{table}[ht]

\caption{Kendall rank correlation coefficients between AOPC and NAOPC$_{\text{beam}}$ rankings across datasets. Coefficients are calculated separately for model rankings and feature attribution method (FA) rankings, showing that normalization impacts the model rankings more than the feature attribution method rankings.}
\label{tab:correlations_all}
\adjustbox{max width=\textwidth}{%
    \centering
    \begin{tabular}{llcc}
    \toprule
     Dataset & Group & Comp & Suff \\
    \midrule
    % \multirow{2}{*}{Yelp$_{\text{short}}$} & Model &  0.81 &  0.00 \\
    % & FA &  0.90  & 0.92 \\
    \multirow{2}{*}{Yelp$_{\text{long}}$} &  Model &  0.87 & 0.47 \\
    & FA &  0.97 &  0.97 \\
    \midrule
    % \multirow{2}{*}{SST-2$_{\text{short}}$} &  Model & 0.92 & 0.67 \\
    % & FA  & 0.87 & 0.93 \\
    \multirow{2}{*}{SST-2$_{\text{long}}$}&  Model  & 0.89 & 0.43 \\
    & FA & 0.90 & 0.92 \\
    \midrule
    \multirow{2}{*}{IMDB$_{\text{long}}$}
    &  Model & 0.93 & 0.72 \\
    &  FA & 0.99 & 0.97 \\
    \midrule
    \multirow{2}{*}{SNLI$_{\text{long}}$}
    &  Model & 0.71 & 1.0 \\
    &  FA & 0.81 & 0.86 \\
    \midrule
    \multirow{2}{*}{AG-News$_{\text{long}}$}
    &  Model & 0.25 & 0.67 \\
    &  FA & 0.90 & 0.83 \\
    \bottomrule
    \end{tabular}
    }
\end{table}
\subsection{NAOPC$_{\text{beam}}$ accurately approximates NAOPC$_{\text{exact}}$}
Our analysis demonstrates that NAOPC$_{\text{beam}}$ accurately approximates NAOPC$_{\text{exact}}$ across various dataset dimensions. For low-dimensional input examples, \Cref{fig:normalization_results_short} shows nearly identical rankings produced by NAOPC$_{\text{beam}}$ and NAOPC$_{\text{exact}}$ on Yelp$_{\text{short}}$. \Cref{fig:sst2_normalized_results} depicts similar results for SST2$_{\text{short}}$. 

% For high-dimensional inputs, we found that increasing the beam size beyond 5 yielded minimal changes on all datasets except for AG-News$_{\text{long}}$. \Cref{fig:increasing_beams_result} depicts these results for RoBERTa$_{\text{Yelp}}$ and BERT$_{\text{Yelp}}$ on Yelp$_{\text{long}}$, and BERT$_{\text{AG-News}}$ on AG-News$_{\text{long}}$. \Cref{fig:all_beam_sizes}, \ref{fig:beam_snli}, and \ref{fig:beam_ag_news} depicts the results for the other models and datasets. We discuss the datasets' different beam sizes in the next section.

\begin{figure}[ht]
    \centering
    \begin{subfigure}{0.23\textwidth}
        \includegraphics[width=\textwidth]{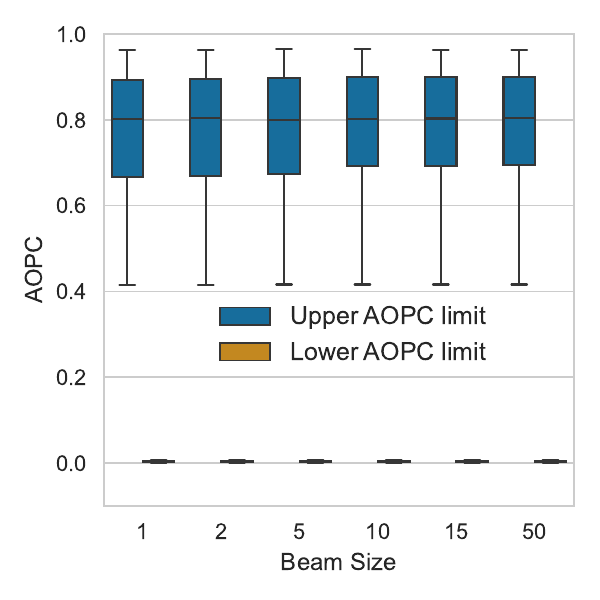}
        \caption{RoBERTa$_{\text{Yelp}}$}
    \end{subfigure}
    \begin{subfigure}{0.23\textwidth}
        \includegraphics[width=\textwidth]{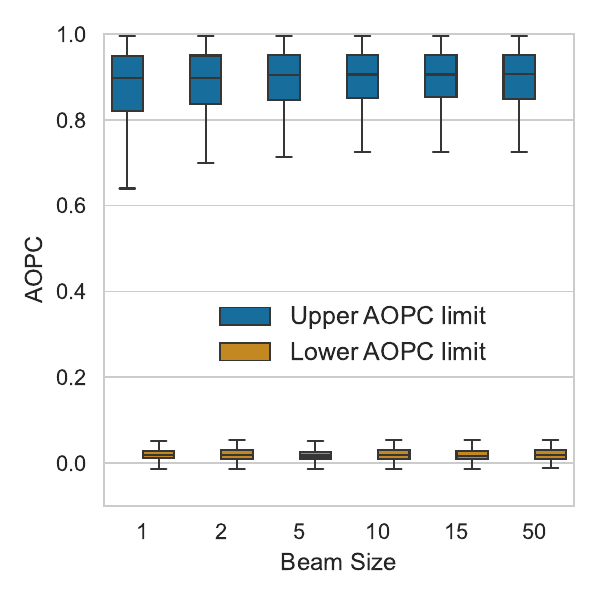}
        \caption{BERT$_{\text{Yelp}}$}
    \end{subfigure}
    \begin{subfigure}{0.46\textwidth}
        \includegraphics[width=\textwidth]{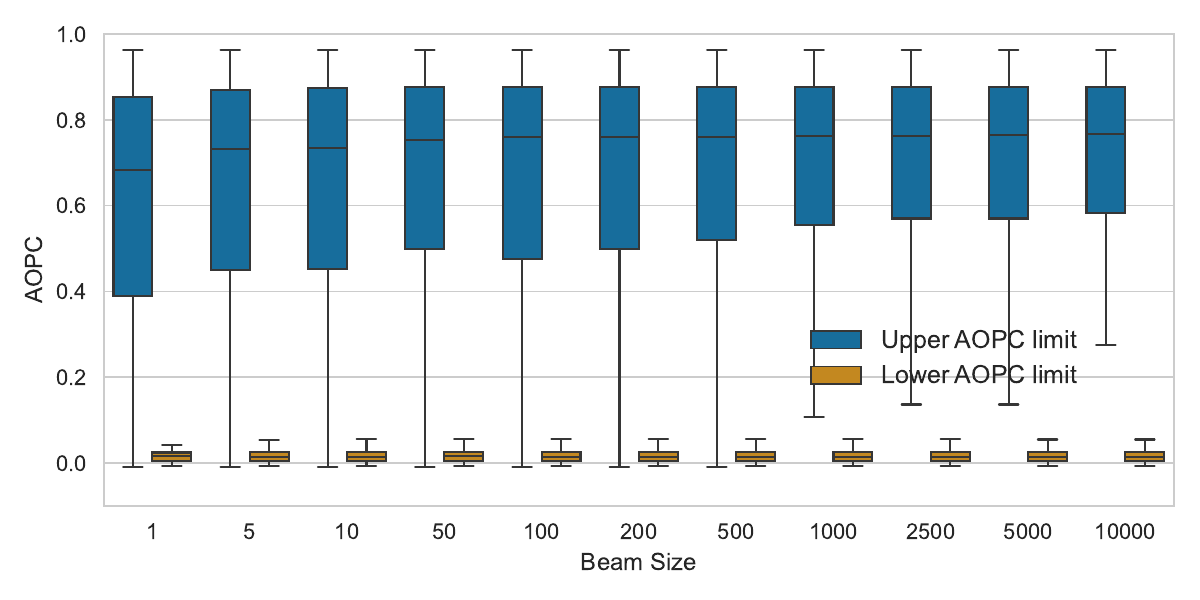}
        \caption{BERT$_{\text{AG-News}}$}
    \end{subfigure}
    \caption{Lower and upper AOPC limits calculated with NAOPC$_{\text{beam}}$ using different beam sizes. RoBERTa$_{\text{Yelp}}$ and BERT$_{\text{Yelp}}$ (a,b) stabilize at $B=5$, while BERT$_{\text{AG-News}}$ (c) requires $B=1000$ for stable results. }
    \label{fig:increasing_beams_result}
\end{figure}

For RoBERTa$_{\text{Yelp}}$ and BERT$_{\text{Yelp}}$ on Yelp$_{\text{long}}$, \Cref{fig:increasing_beams_result} shows that a beam size of 5 is sufficient for stable results. However, the same figure demonstrates that BERT$_{\text{AG-News}}$ requires a substantially larger beam size. \Cref{fig:all_beam_sizes,fig:beam_snli,fig:beam_ag_news} confirm this pattern across all datasets, with AG-News being the only dataset requiring a larger beam size. We explore the reasons for this behavior and its implications in the next section.

\section{Discussion}

\subsection{Does NAOPC require too much compute to be practically useful?}\label{sec:dis_beamsize}
NAOPC is computationally more intensive than AOPC. Computing AOPC requires $N$ forward passes, NAOPC$_{\text{beam}}$ requires $BN^2$ forward passes, and NAOPC$_{\text{exact}}$ requires $N!$ forward passes, where $N$ is the number of input features, and $B$ is the beam size. While the exponential complexity of NAOPC$_{\text{exact}}$  makes it impractical for most inputs, our experiments demonstrate that NAOPC$_{\text{beam}}$ is feasible in many scenarios.

The computational cost of NAOPC$_{\text{beam}}$ depends on beam size, input length, and model inference time. With a small beam size (B=5), which suffices for many datasets, computing NAOPC$_{\text{beam}}$ for BERT (110 million parameters) on a hundred-feature example took around one minute on an A100 GPU. While requirements grow quadratically with input length, we found this manageable for several hundred tokens: processing 512-token inputs (BERT's maximum) took approximately 10 minutes per example, averaged over 100 examples. Since NAOPC$_{\text{beam}}$ scales linearly with model size, larger models remain feasible to evaluate. Importantly, normalization factors only need to be computed once per model-dataset pair and can be reused, significantly reducing overall cost.

However, some datasets require larger beam sizes for accurate normalization. For instance, models trained on AG-News required a beam size of 1,000 to achieve stable results. This requirement does not appear to relate to input length, as AG-News comprises shorter sequences than Yelp and SST2. We speculate this might be due to feature interactions where multiple features must be removed together to measure their true impact on the model's prediction. In such cases, a larger beam size is necessary to ensure these feature combinations are explored during the search. However, further research is needed to verify this hypothesis and understand what drives beam size requirements.

To help researchers assess requirements upfront, we provide tools for estimating necessary beam sizes for specific model-dataset combinations, allowing evaluation of cross-model comparison feasibility given computational constraints.

\subsection{Should one always normalize the AOPC scores?}
Given that computing, NAOPC requires additional computational resources, a natural question arises: when is this extra computation necessary? Our findings demonstrate that normalization is essential in two scenarios: comparing AOPC scores across different models and interpreting individual scores in relation to a model's theoretical limits.

For cross-model comparisons, normalization is necessary even when comparing models with identical architectures trained on the same dataset but with different random seeds, as even slight variations in model parameters can lead to different AOPC limits. Without normalization, a score of 0.25 could be near-optimal for one model but mediocre for another, making cross-model comparisons misleading. Therefore, we recommend normalizing AOPC scores in all cases except when only comparing the relative ranking of feature attribution methods within a single model, where the absolute values of the scores are not relevant. The fundamental importance of normalization for valid cross-model comparisons calls into question previous research findings. Studies that compared unnormalized AOPC scores across different models may need to be re-evaluated.

\subsection{Why are some models more faithful than others?}
By normalizing AOPC scores, we can now meaningfully compare explanation faithfulness across models and interpret how far they are from optimal performance. Our analysis reveals substantial differences in faithfulness even after normalization. For instance, on Yelp$_{\text{long}}$, all explanation methods achieved substantially higher comprehensiveness for BERT$_{\text{IMDB}}$ than RoBERTa$_{\text{IMDB}}$. For DecompX, the best method, the difference was 0.75 and 0.44. Why do the explanation methods produce less faithful explanations for RoBERTa$_{\text{IMDB}}$? 

We hypothesize these variations stem from differences in how closely models align with the feature attribution methods' assumptions. Most feature attribution methods rely on simplified assumptions about models' inner mechanisms, such as feature independence~\cite{bilodeau2024impossibility}. RoBERTa$_{\text{IMDB}}$ consistently achieves low comprehensiveness scores across all tested attribution methods, even after normalization, suggesting a fundamental mismatch between how this model processes information and current attribution methods' assumptions. However, further research examining the models' internal mechanisms would be needed to verify this hypothesis.

 \subsection{Why does normalization have a bigger impact on shorter text datasets?}
 \Cref{fig:normalization_results_long} and \Cref{fig:normalization_results_short} show that normalization changes the AOPC results more on shorter sequences than on longer. What causes this difference?  We hypothesize that models typically rely on a small subset of features when processing long inputs. When measuring AOPC (comprehensiveness and sufficiency), we remove tokens based on their importance sequentially. In long texts, most removals have no effect since they were irrelevant in the model's decision. When averaging across all removal steps, these zero-effect steps dilute the AOPC scores---bringing sufficiency scores closer to zero and comprehensiveness scores closer to one---making the impact of normalization less visible since both the raw scores and their theoretical limits converge. If there are fewer unused tokens in shorter texts, then fewer zero-effect steps are included in the average. This makes normalization effects more pronounced. If we could somehow consider only the tokens impacting the model’s decision, we hypothesize that the normalization effect would be similar on short and long inputs. 

\section{Related Work}

Researchers have raised several criticisms against AOPC and other perturbation-based faithfulness metrics, which fall into three main categories. First, perturbing inputs can create out-of-distribution examples, potentially conflating distribution shifts with feature importance~\cite{ancona2017towards, hookerBenchmarkInterpretabilityMethods2019, haseOutofDistributionProblemExplainability2021}. Second, perturbations often yield inputs that appear non-sensical to humans, though this should not affect faithfulness evaluation~\cite{feng2018pathologies, bastingsElephantInterpretabilityRoom2020a, jacoviFaithfullyInterpretableNLP2020}. Third, these metrics can be viewed as attribution methods themselves, potentially measuring similarity between methods rather than true faithfulness~\cite{zhou2022solvability, juLogicTrapsEvaluating2023}.

Our work focuses on sufficiency and comprehensiveness due to their widespread use in cross-model comparisons~\cite{bhallaDiscriminativeFeatureAttributions2023,liFaithfulExplanationsText2023, chrysostomouEmpiricalStudyExplanations2022, liuImproveInterpretabilityNeural2022}. However, researchers have also developed alternative faithfulness metrics to address the potential limitations of AOPC. Decision-flip metrics track when model predictions change as features are removed~\cite{chrysostomouEmpiricalStudyExplanations2022}. Monotonicity and the faitfulness correlation metric (CORR) measure whether higher attribution scores correspond to larger changes in model output~\cite{arya2019one}. Sensitivity-n tests if the sum of attribution scores equals the total change in model output when removing the  features~\cite{ancona2017towards}. 

While these alternatives were designed to provide different perspectives on faithfulness, we suspect that they share some of AOPC's fundamental limitations. Like AOPC, decision-flip metrics may produce misleading results when comparing models that rely on different numbers of features because fewer features need to be removed to significantly change the model's prediction, resulting in artificially better scores. Similarly, we expect metrics like sensitivity-n, monotonicity, and CORR to struggle with feature interactions because they assume attribution scores can be assigned independently. This assumption likely breaks down when feature importance depends on feature interactions rather than independent features. For example, consider a model using OR operations ($x_1 \lor x_2$). Sensitivity-n requires attribution scores to sum to the total change when both features are removed ($e_1 + e_2 = 1$) but also requires each score to equal its individual impact ($e_1=e_2=0$), creating an impossible mathematical constraint. These limitations suggest a broader need to develop faithfulness metrics that can account for model-specific characteristics and complex feature interactions.

\section{Conclusion}
Our study exposes critical weaknesses in current faithfulness evaluation practices for feature attribution methods. Using simple toy models, we demonstrated how models' inner mechanisms significantly influence AOPC's lower and upper limits, potentially leading to misleading cross-model comparisons. Moreover, without knowing these limits, it becomes difficult to interpret AOPC scores effectively. These findings challenge the validity of conclusions drawn from cross-model AOPC score comparisons in many influential studies. To address these issues, we introduced NAOPC, a normalized measure that mitigates model-dependent bias while preserving the ability to compare feature attribution methods within individual models. NAOPC enables accurate evaluation of feature attribution faithfulness across different models, advancing the field towards more robust explanation assessment. While NAOPC addresses these fundamental issues, its computational complexity suggests the need for future research into faster interpretable faithfulness metrics that maintain cross-model comparability.

\section*{Limitations}
Our findings indicate that normalization did not alter the faithfulness ranking of feature attribution methods within a model. This suggests that normalization is unnecessary when comparing AOPC scores produced using one model and one dataset. Nonetheless, our evaluation did not cover a sufficient variety of models, tasks, and datasets to rule out the necessity of normalization for certain within-model comparisons. We leave the evaluation of more models, datasets, and tasks to future work.

In addition, as discussed in \Cref{sec:dis_beamsize}, with its $O(BN^2)$ time complexity, NAOPC$_{\text{beam}}$ will be prohibitively slow for certain datasets, especially for those requiring large beam sizes. Most of our datasets and models required small beam sizes (B=5), but AG-News$_{\text{long}}$ required a large beam size (B=1000). However, it is better with a slow evaluation than an inaccurate one. Moreover, we provide software tools to help researchers determine the necessary beam size for their specific use case. This allows researchers to assess the computational requirements beforehand and plan their experiments accordingly, deciding whether cross-model comparisons are feasible for their dataset and computational budget.

\section*{Ethical considerations}
The ability to explain deep neural network decisions is crucial for ensuring their responsible deployment, particularly in high-stakes domains such as healthcare, legal systems, and financial services. When a diagnostic model suggests treatment or when a neural network influences a parole decision, stakeholders must be able to scrutinize and validate the reasoning behind these recommendations. However, explanations are only valuable if they faithfully represent the model's decision-making process.

Our work reveals that current methods for evaluating explanation faithfulness can be misleading, potentially giving false confidence in explanation methods that do not accurately reflect model behavior. This is particularly concerning because unreliable explanations might lead to unwarranted trust in neural networks or mask potential biases in their decision-making processes. For instance, \citet{kayserFoolMeOnce2024} demonstrated that incorrect explanations can persuade phycisians into an incorrect diagnosis. By providing a more reliable evaluation framework through NAOPC, we contribute to the development of more trustworthy explanation methods, ultimately supporting the responsible deployment of deep neural networks in society.
\section*{Acknowledgments}
This research was partially funded by the Innovation Fund Denmark via the Industrial Ph.D. Program (grant no. 2050-00040B) and Academy of Finland (grant no. 322653). We thank Simon Flachs, Nina Frederikke Jeppesen Edin, and Victor Petrén Bach Hansen for revisions.

% Bibliography entries for the entire Anthology, followed by custom entries
%\bibliography{anthology,custom}
% Custom bibliography entries only
\bibliography{my_lib, custom}

\appendix

\section{Model Details and Access}\label{app:a}

In \Cref{tab:apdx_models}, we present an overview of the six models used in our study. For each model, we provide details on the architecture, training data, and a direct link to the corresponding pre-trained weights available on HuggingFace.

\begin{table*}[ht]
    \caption{Overview of the public models from Hugging Face used in this paper.}
    \label{tab:apdx_models}
    \adjustbox{max width=\textwidth}{%
    \centering
    \begin{tabular}{clll}
    \toprule
        Model & Architecture (Param) & Training data & HuggingFace link\\
        \midrule
        BERT$_{\text{Yelp}}$& BERT (110M) & Yelp & textattack/bert-base-uncased-yelp-polarity\\
        RoBERTa$_{\text{Yelp}}$ & RoBERTa (110M) & Yelp & VictorSanh/roberta-base-finetuned-yelp-polarity\\
        BERT$_{\text{IMDB}}$ & BERT (110M)& IMBD & textattack/bert-base-uncased-imdb\\
        RoBERTa$_{\text{IMDB}}$ & RoBERTa (110M)& IMBD & textattack/roberta-base-imdb\\
        BERT$_{\text{SST2}}$ & BERT (110M)& SST2 & textattack/bert-base-uncased-SST-2\\
        RoBERTa$_{\text{SST2}}$ & RoBERTa(110M) & SST2 & textattack/roberta-base-SST-2\\
        BERT$_{\text{AG-News}}$ & BERT (110M)& AG-News & textattack/bert-base-uncased-ag-news\\
        RoBERTa$_{\text{AG-News}}$ & RoBERTa (110M)& AG-News & textattack/roberta-base-ag-news\\
        DistilBERT$_{\text{AG-News}}$ & DistilBERT (66M)& AG-News & textattack/distilbert-base-uncased-ag-news\\
        BERT$_{\text{SNLI}}$ & BERT (110M)& SNLI & textattack/bert-base-uncased-snli\\
        DistilBERT$_{\text{SNLI}}$ & DistilBERT (66M)& SNLI & textattack/distilbert-base-cased-snli\\
        GPT-2$_{\text{SNLI}}$ & GPT-2 (124M)& SNLI & varun-v-rao/gpt2-snli-model1\\
        \bottomrule
    \end{tabular}}
\end{table*}

\section{Analysis of AOPC Score Limit Variability Across Models on the SST2 Dataset}\label{app:b}

\begin{figure*}[h]
    \centering
    \begin{subfigure}{0.4\textwidth}
        \includegraphics[width=\linewidth]
        {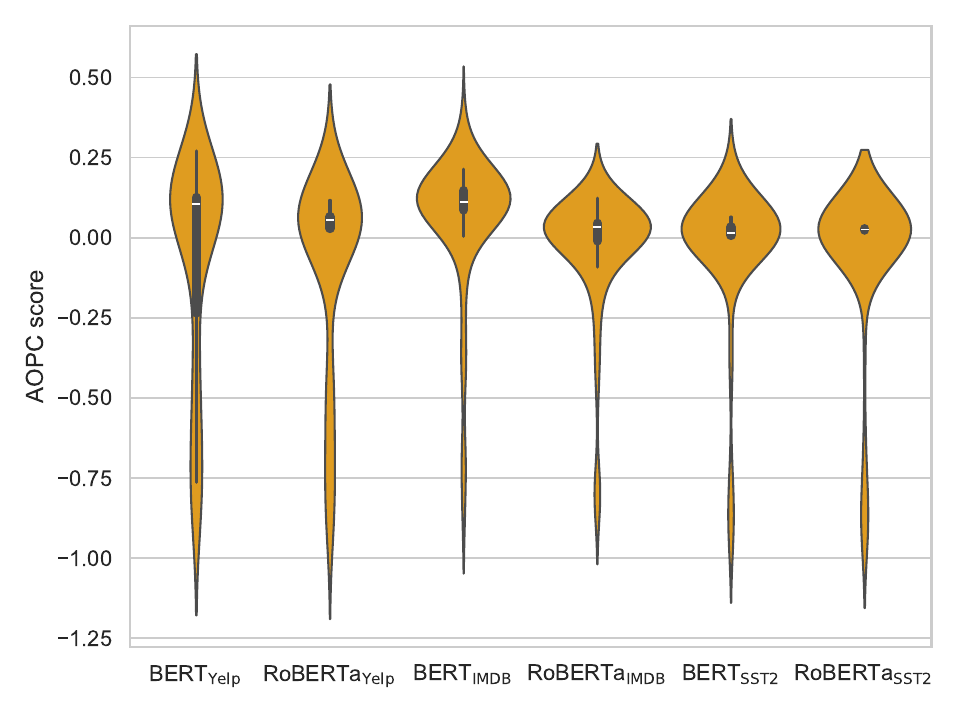}
        \caption{Lower AOPC limit}
    \end{subfigure}
    \begin{subfigure}{0.4\textwidth}
        \includegraphics[width=\linewidth]{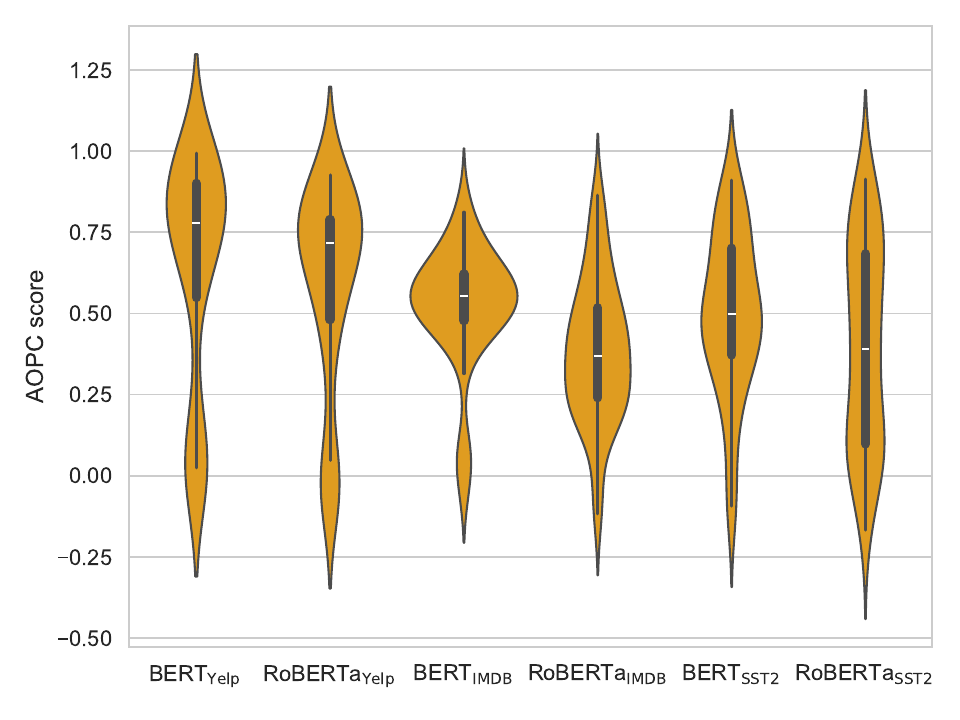}
        \caption{Upper AOPC limit}
    \end{subfigure}
    \caption{Distributions of lower and upper AOPC limits for various models on the SST2$_{\text{short}}$ test set. Each distribution reflects the range of possible AOPC scores for a given model, influenced by individual input examples. The inter-model variations demonstrate the need for normalization when comparing AOPC scores across different models.}
    \label{fig:sst2-upper-lower}
\end{figure*}

\Cref{fig:sst2-upper-lower} shows the distributions of lower and upper AOPC score limits for different models on the SST2$_{\text{short}}$ test set. The presence of distributions rather than single values for each model highlights the influence of individual inputs on AOPC limits. The notable differences in these distributions across models underscore the importance of normalization when comparing comprehensiveness and sufficiency scores between models.

\section{NAOPC Comparison on SST2$_{\text{short}}$}\label{app:c}

\begin{figure*}[ht]
    \centering
    \begin{subfigure}{0.4\textwidth}
        \includegraphics[width=\linewidth]
        {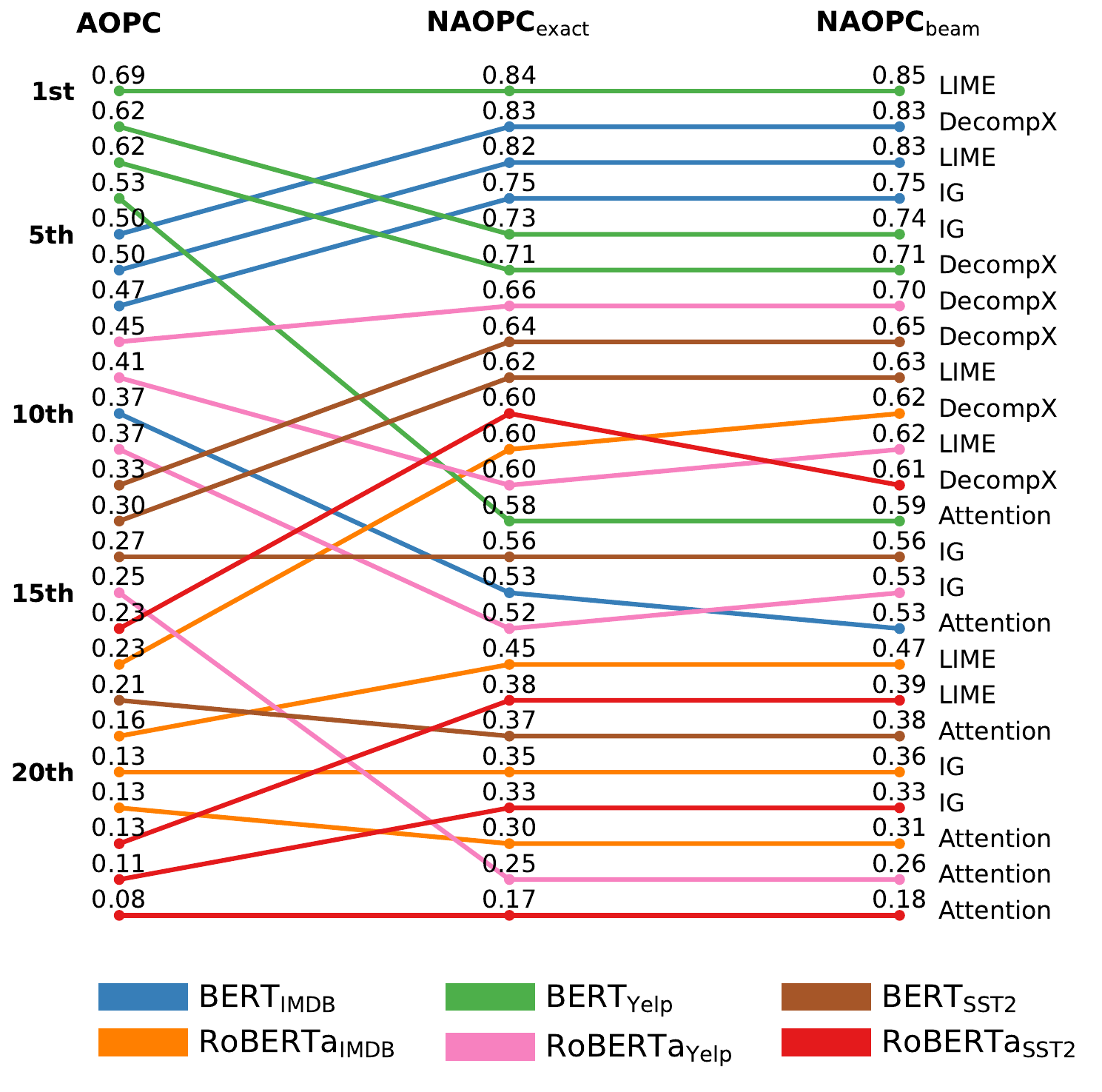}
        \caption{Comprehensiveness Order}
    \end{subfigure}
    \begin{subfigure}{0.4\textwidth}
        \includegraphics[width=\linewidth]{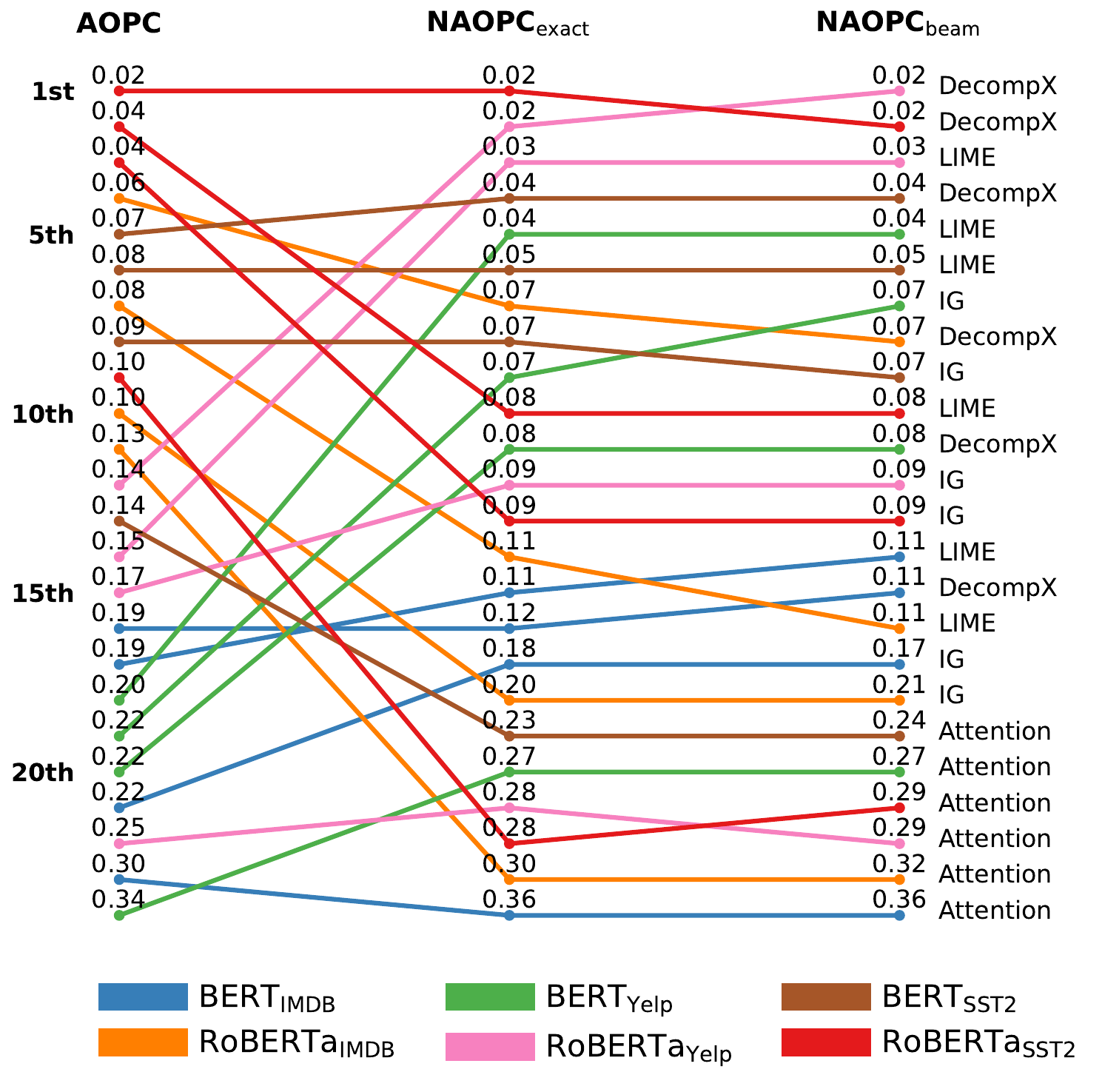}
        \caption{Sufficiency Order}
    \end{subfigure}

    \caption{The difference in rankings when normalizing on the SST-2 dataset}
    \label{fig:sst2_normalized_results}
\end{figure*}

This section presents a detailed comparison of NAOPC$_{\text{beam}}$ and NAOPC$_{\text{exact}}$ on the SST2$_{\text{short}}$ dataset. \Cref{fig:sst2_normalized_results} illustrates the rankings produced by both methods when evaluating the comprehensiveness and sufficiency of the dataset. The figure shows almost identical rankings produced by NAOPC$_{\text{beam}}$ and NAOPC$_{\text{exact}}$.

\section{NAOPC$_{\text{beam}}$ results on AG-News$_{\text{long}}$ and SNLI$_{\text{long}}$}

\begin{figure*}[ht]
    \centering
    \begin{subfigure}{0.4\textwidth}
        \includegraphics[width=\textwidth]{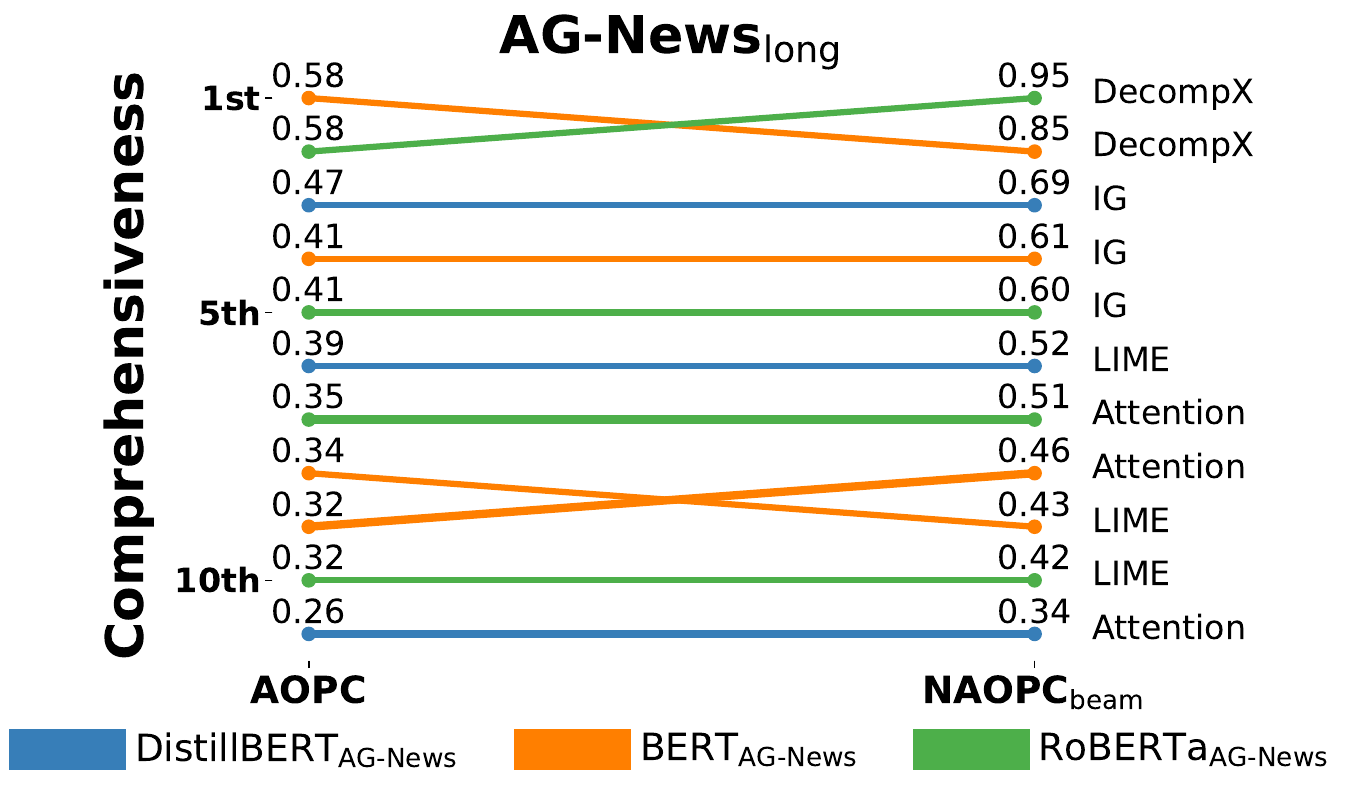}
    \end{subfigure}
    \begin{subfigure}{0.35\textwidth}
        \includegraphics[width=\textwidth]{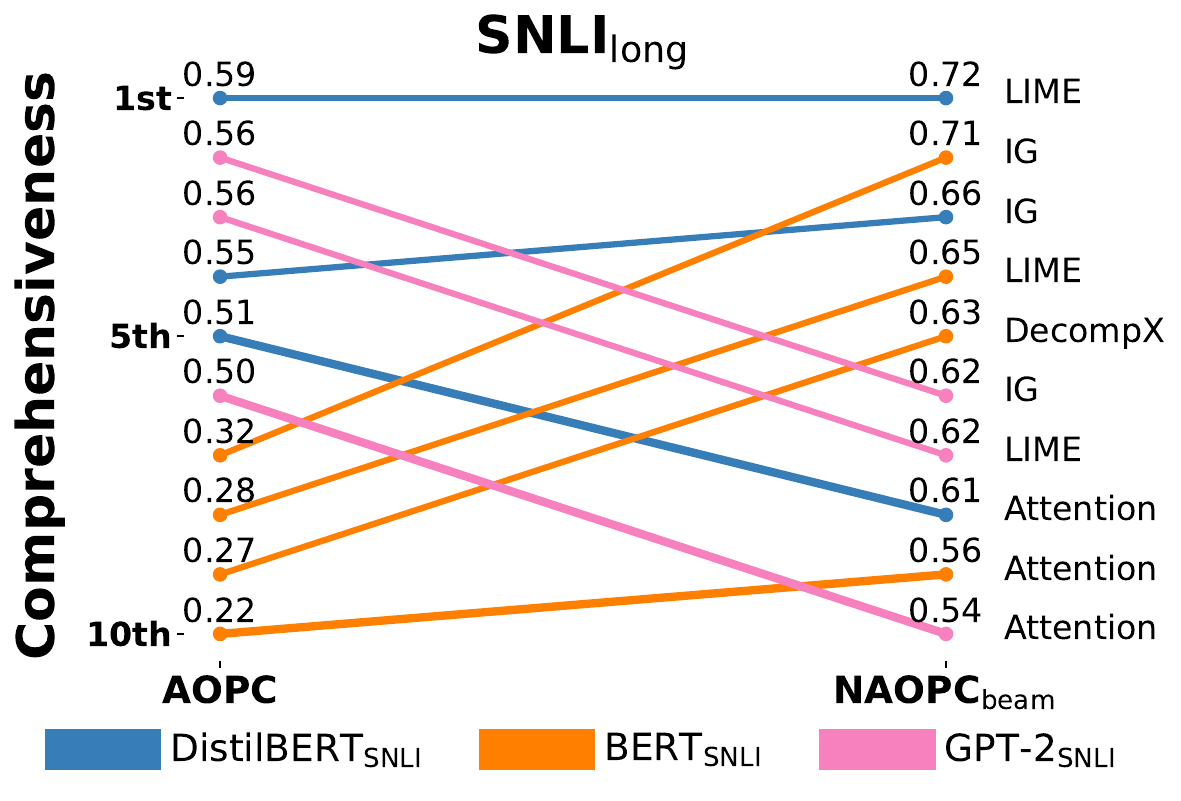}
    \end{subfigure}
    \begin{subfigure}{0.4\textwidth}
        \includegraphics[width=\textwidth]{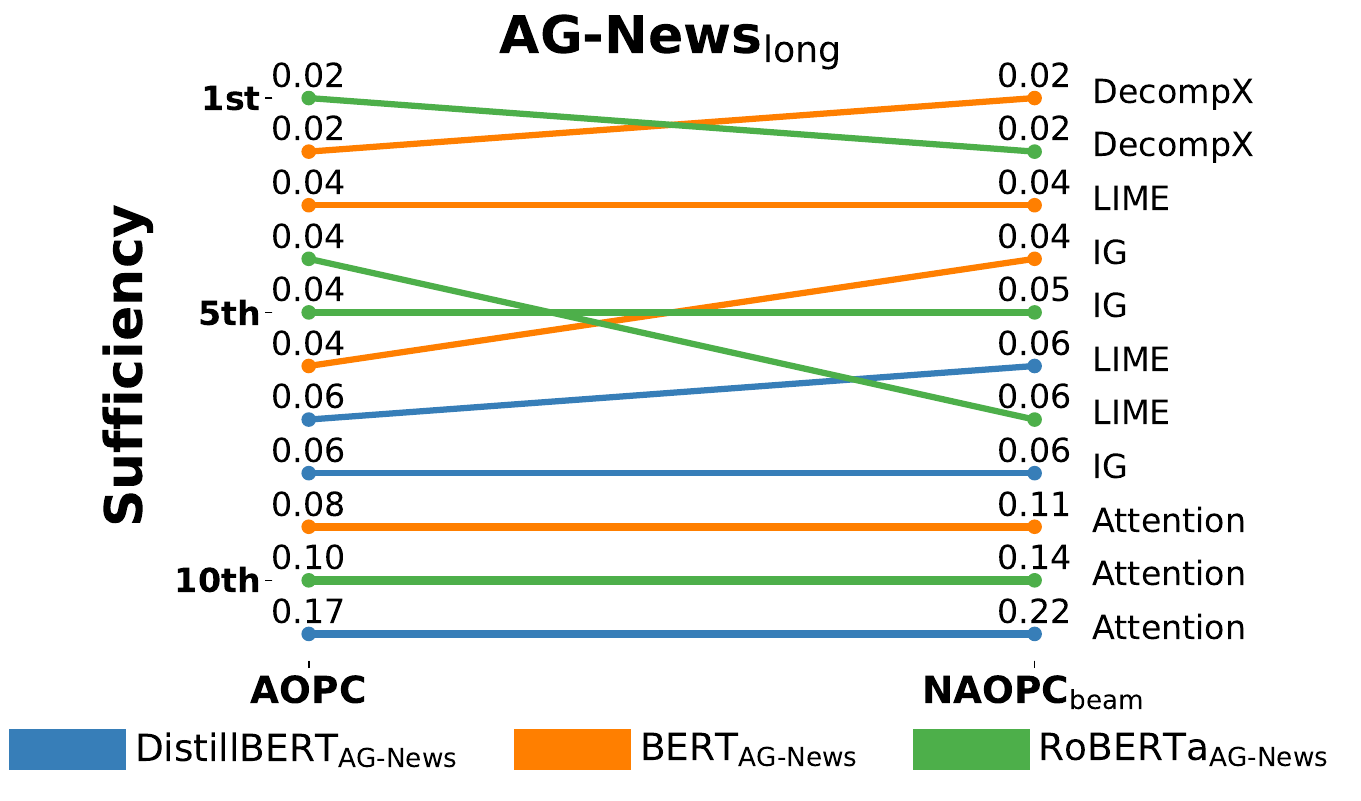}
    \end{subfigure}
    \begin{subfigure}{0.35\textwidth}
        \includegraphics[width=\textwidth]{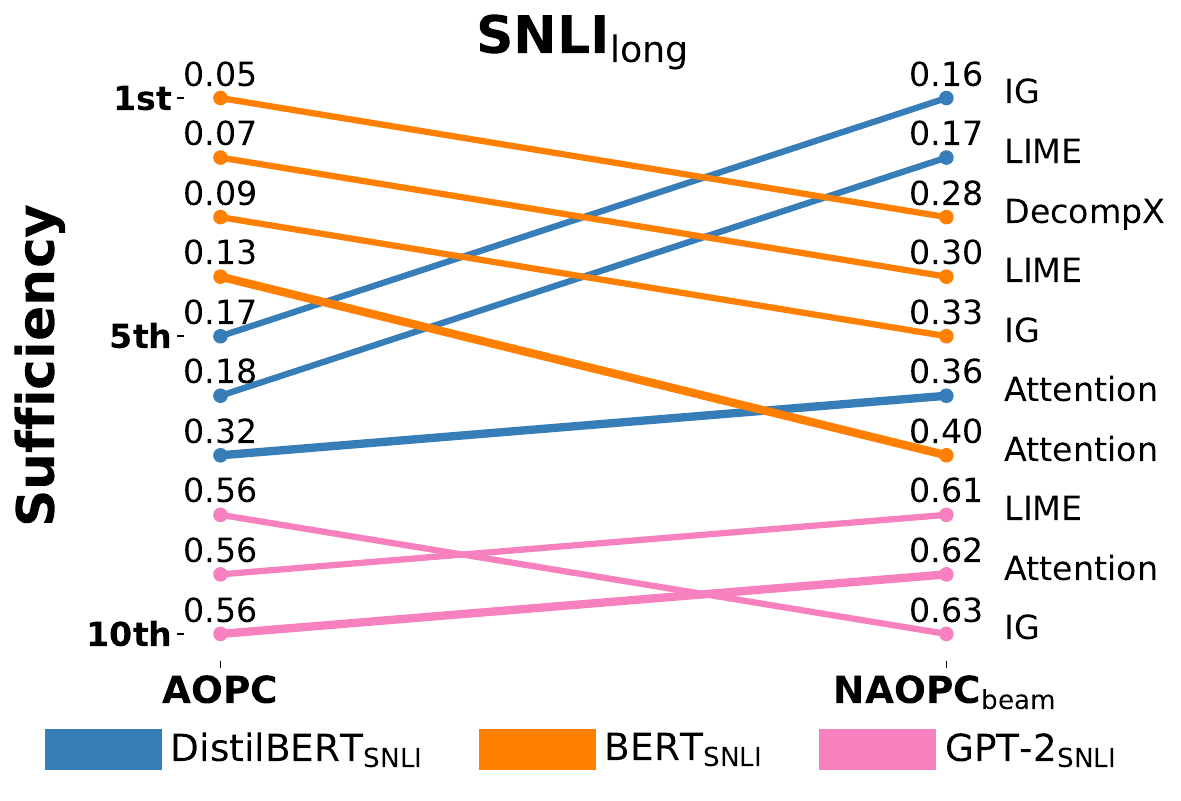}
    \end{subfigure}

    \caption{Faithfulness ranking of model and feature attribution method pairs when evaluated using AOPC and NAOPC$_{\text{beam}}$ on AG-News and SNLI. }
    \label{fig:normalization_results_agnews_snli}
\end{figure*}

\Cref{fig:normalization_results_agnews_snli} depicts the difference between AOPC and NAOPC$_{\text{beam}}$ on AG-News$_{\text{long}}$ and SNLI$_{\text{long}}$. We see similar results as in \Cref{fig:normalization_results_long}. However, on SNLI$_{\text{long}}$, LIME and IG seem to be better than DecompX. Also, GPT-2's sufficiency and comprehensiveness scores are similar. We speculate this is because we perturb by replacing with the end-of-sequence token (GPT-2 does not support mask tokens nor pad tokens). End-of-sequence tokens in the middle of a sentence may quickly make the input out-of-distribution, therefore changing the model's output, even when perturbing unimportant features.

\section{Impact of Beam Size on NAOPC$_{\text{beam}}$ Across Various Datasets}
\label{app:d}

\begin{figure*}[ht]
    \centering
    \includegraphics[width=0.8\textwidth]{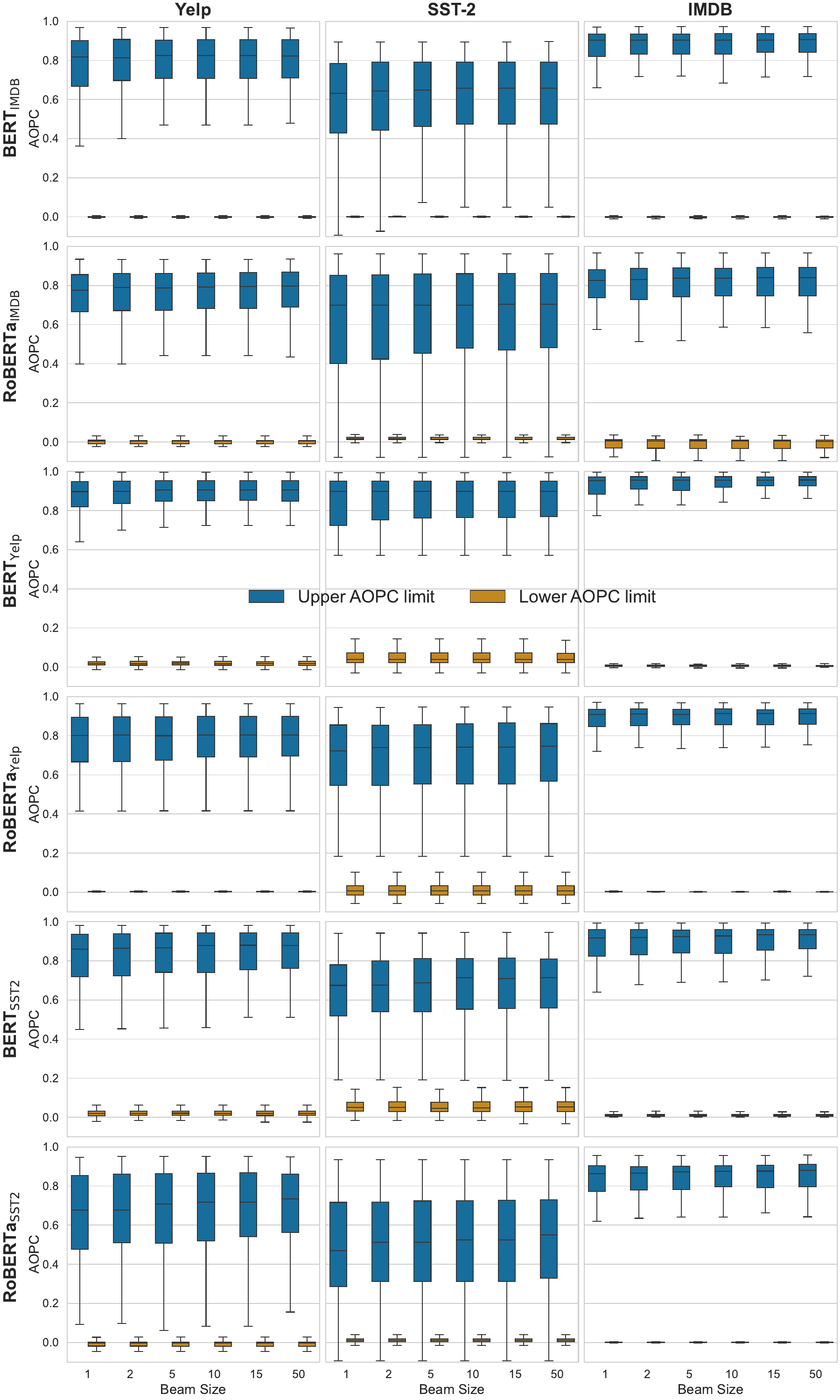}

    \caption{Boxplots showing the distribution of NAOPC$_{\text{beam}}$ values across different beam sizes for various models and datasets.}.
    \label{fig:all_beam_sizes}
\end{figure*}

As illustrated in \Cref{fig:all_beam_sizes}, the relationship between increasing beam size and NAOPC$_{\text{beam}}$ is examined across various models and datasets. The findings indicate that while an initial expansion of beam size results in variability in the upper and lower bounds, further increases beyond a beam size of 5 lead to a convergence trend. This pattern is consistently observed across different models and datasets, particularly evident in the results for RoBERTa$_{\text{Yelp}}$ and BERT$_{\text{Yelp}}$ on the Yelp$_{\text{long}}$ dataset.

\begin{figure*}[h]
    \centering
    \begin{subfigure}{0.32\textwidth}
        \includegraphics[width=\linewidth]
        {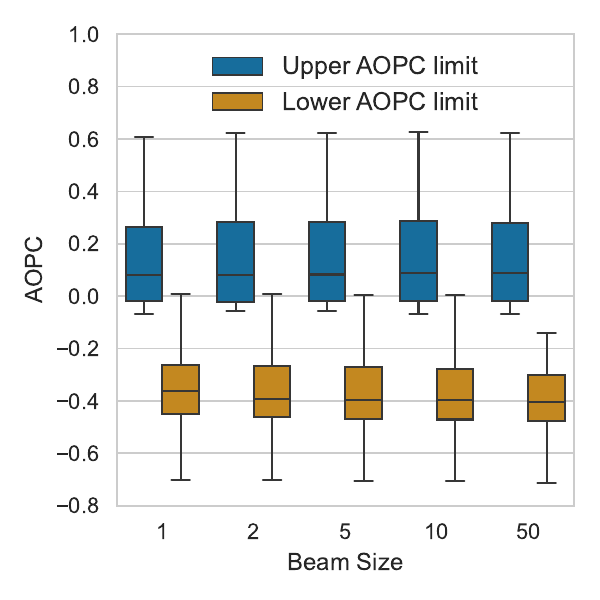}
        \caption{BERT}
    \end{subfigure}
    \begin{subfigure}{0.32\textwidth}
        \includegraphics[width=\linewidth]{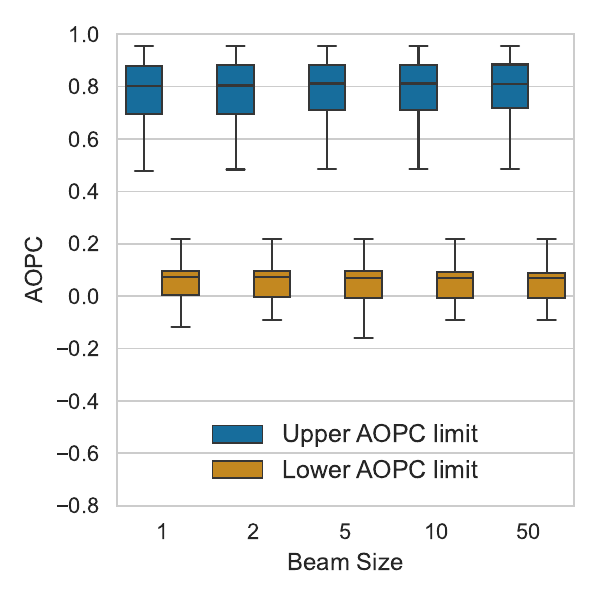}
        \caption{DistilBERT}
    \end{subfigure}
    \begin{subfigure}{0.32\textwidth}
        \includegraphics[width=\linewidth]{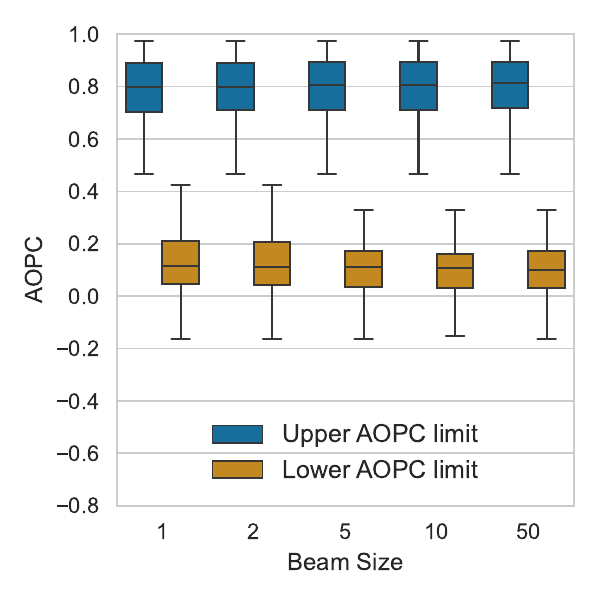}
        \caption{GPT-2}
    \end{subfigure}
    \caption{Boxplots showing the distribution of NAOPC$_{\text{beam}}$ values across different beam sizes for SNLI.}
    \label{fig:beam_snli}
\end{figure*}

\begin{figure*}[h]
    \centering
    \begin{subfigure}{0.6\textwidth}
        \includegraphics[width=\linewidth]
        {figures/boxplots/ag_news_bert-base-uncased-ag-news_increasing_beam_sizes.pdf}
        \caption{BERT}
    \end{subfigure}
    \begin{subfigure}{0.6\textwidth}
        \includegraphics[width=\linewidth]{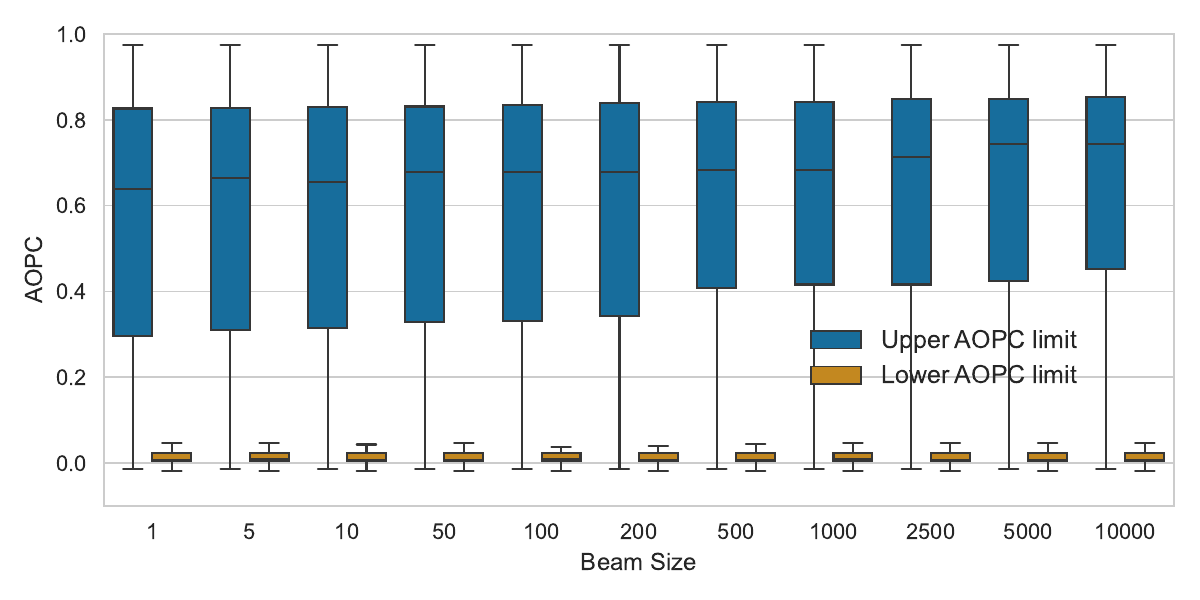}
        \caption{RoBERTa}
    \end{subfigure}
    \begin{subfigure}{0.6\textwidth}
        \includegraphics[width=\linewidth]{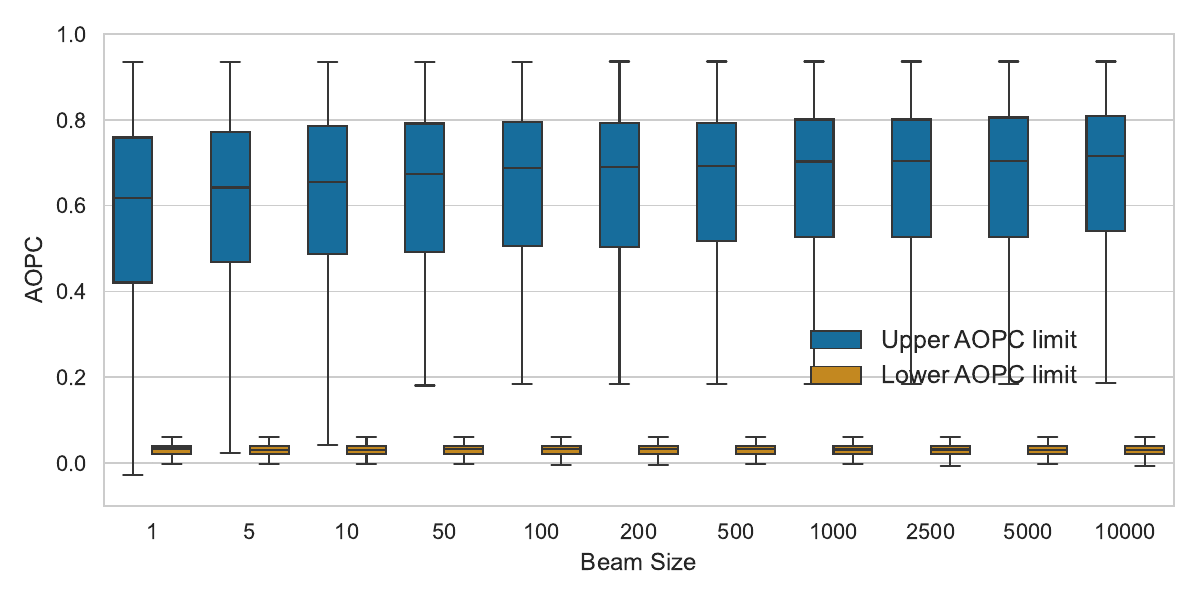}
        \caption{DistilBERT}
    \end{subfigure}
    \caption{Boxplots showing the distribution of NAOPC$_{\text{beam}}$ values across different beam sizes for AG-News.}
    \label{fig:beam_ag_news}
\end{figure*}

\section{Licences}
SNLI uses a cc-by-sa-4.0 license. AG-News does not have a specific license, but the authors state that it should only be used for non-commercial purposes\footnote{\url{http://groups.di.unipi.it/~gulli/AG_corpus_of_news_articles.html}}. SST-2 and IMDB are both created by StanfordNLP, who do not specify a license, but writes that you must cite their papers if using the dataset\footnote{\url{https://ai.stanford.edu/~amaas/data/sentiment/}}\footnote{\url{https://nlp.stanford.edu/sentiment/}}. We used Yelp from Huggingface\footnote{\url{https://huggingface.co/datasets/fancyzhx/yelp_polarity}}. The original webpage with the Yelp dataset and License no longer exists. Considering that thousands of other papers use this dataset, it is most likely okay to use, but we cannot guarantee it since we could not find its license. The language models from Huggingface use an MIT license.

\end{document}